\documentclass{article}

\usepackage[utf8]{inputenc}
\usepackage{url}
\usepackage{hyperref}
\usepackage{graphicx}
\usepackage[dvipsnames,table]{xcolor}
\usepackage{enumitem}
\usepackage{booktabs}
\usepackage{subcaption}
\usepackage{mathptmx}
\usepackage{pifont}
\usepackage{multirow}
\usepackage{float}
\usepackage[toc,page]{appendix}
\usepackage{longtable}

\usepackage[russian,english]{babel}
\definecolor{darkgreen}{rgb}{0.0, 0.2, 0.13}

\title{Beyond Plain Toxic: Detection of Inappropriate Statements on Flammable Topics  for the Russian Language}
\author{Nikolay Babakov \and Varvara Logacheva \and Alexander Panchenko \\ 
\{n.babakov, v.logacheva, a.panchenko\}@skoltech.ru \\
Skolkovo Institute of Science and Technology, Moscow, Russia}

\date{}

\begin{document}

\maketitle

\begin{abstract}
Toxicity on the Internet, such as hate speech, offenses towards particular users or groups of people, or the use of obscene words, is an acknowledged problem. However, there also exist other types of \textit{inappropriate} messages which are usually not viewed as toxic, e.g. as they do not contain explicit offences. Such messages can contain covered toxicity or generalizations, incite harmful actions (crime, suicide, drug use), provoke ``heated'' discussions. Such messages are often related to particular \textit{sensitive topics}, e.g. on politics, sexual minorities, social injustice which more often than other topics, e.g. cars or computing, yield toxic emotional reactions. At the same time, clearly not all messages within such flammable topics are inappropriate.

A machinery able to detect inappropriateness in a conversation could be useful for: (i) making the communication on the Internet safer, more productive and inclusive by flagging truly inappropriate content passing toxicity filters while not banning messages blindly topically; (ii) for detection of inappropriate messages generated by automatic systems, e.g. neural chatbots, due to bias in the training data; (iii) for filtering training data from bias for language models (e.g. BERT and GPT-2).

Towards this end, in this work, we present two text collections labelled according to binary notion of \textit{inapropriateness} and a multinomial notion of \textit{sensitive topic}. Assuming that the notion of inappropriateness is common among people of the same culture, we base our approach on human intuitive understanding of what is not acceptable and harmful. To objectivise the notion of inappropriateness, we define it in a data-driven way though crowdsourcing. Namely we run a large-scale annotation study asking workers if a given chatbot textual statement could harm reputation of a company created it. Acceptably high values of inter-annotator agreement suggest that the notion of inappropriateness exists and can be uniformly understood by different people. To define the notion of sensitive topics in an objective way we use on guidelines suggested commonly by specialists  of legal and PR department of a large public company as potentially harmful. 

\end{abstract}

\section{Introduction}
\label{intro}

Toxicity is a serious problem in online communication. It can take multiple forms --- from disrespectful hints to actions that transfer from online space to the real world, threatening people's well-being and even lives. Thus, it is crucial to prevent such kinds of abuse in textual form. First, toxicity-free online communication is more pleasant and much more productive. Second, admission of online toxicity implies that it is also acceptable offline, so not stopping it early can have serious consequences. 

Therefore, toxicity detection has recently become a topic of active research in Natural Language Processing (NLP)~\cite{jigsaw_toxic,jigsaw_bias,jigsaw_multi,davidson2017automated,breitfeller-etal-2019-finding}. There exist a lot of works on recognising toxicity in textual communication and also on its prevention by recognising its premises. Some of these techniques have already been implemented in commercial companies. For example, Instagram \href{https://about.instagram.com/blog/announcements/national-bullying-prevention-month}{checks user comments for toxicity before posting}, and if they are toxic, suggests reformulating them before sending.

Another not so obvious, yet serious danger of the spread of toxicity is the fact that large language models, such as GPT-2~\cite{gpt2}, are trained on the large-scale user-generated data from the Internet. If this data contains samples of toxicity, generation models can also learn to produce toxic texts. One well-known case is \href{https://www.theverge.com/2016/3/24/11297050/tay-microsoft-chatbot-racist}{Microsoft Tay chatbot} which was shut down because it started producing racist, sexist, and other offensive tweets after having been fine-tuned on user data. There are numerous analogous failures by dialogue assistants in different languages. A Korean chatbot Luda Lee \href{https://www.straitstimes.com/asia/east-asia/controversy-over-ai-chatbot-in-south-korea-raises-questions-about-ethics-data}{insulted sexual minorities} and people with disabilities. Alice chatbot by Yandex (a major Russian search engine) \href{https://techcrunch.com/2017/10/24/another-ai-chatbot-shown-spouting-offensive-views/?guccounter=1&guce_referrer=aHR0cHM6Ly93d3cuZ29vZ2xlLmNvbS8&guce_referrer_sig=AQAAADx3l6Et-kpX507QbRFcsgjr3if1Y9aQ-gIi7E8AltwElhokAnQaLBD6OdEg1IYvEaZZZfkmwN_HbJ6mOGfaJwP51OqgnAdzY3yxHNde3FaFfrtBzgZP5eAphIUyrxsVFUu4nqFVnqT6P5QlIGRaBFnJYrhsLC64-U_152xrPnpU}{expressed extremist views}, approved violence and suicide. Oleg chatbot by the Russian bank Tinkoff \href{https://meduza.io/news/2019/11/26/chat-bot-tinkoff-banka-posovetoval-klientke-otrezal-paltsy-v-banke-eto-ob-yasnili-obucheniem-na-otkrytyh-dannyh}{advised} a customer to have her fingers cut off. Similarly to Tay, all these chatbots were trained on large amounts of user conversations, tweets, or other user-generated texts. 
Thus, toxicity should be detected not only to make human-to-human communication safer but also not to transmit harmful human biases to machines. In the case of human-to-machine communication, toxicity produced by a machine can cause reputational loss to a company that created it. This is an additional motivation for research groups from the industry to work on this problem.

While toxicity attracts attention of many researchers, it is not the only type of undesired content which can harm human-to-human conversations and distort the training data for language models.

\begin{figure}[ht!]
\centering
\includegraphics[width=0.7\linewidth]{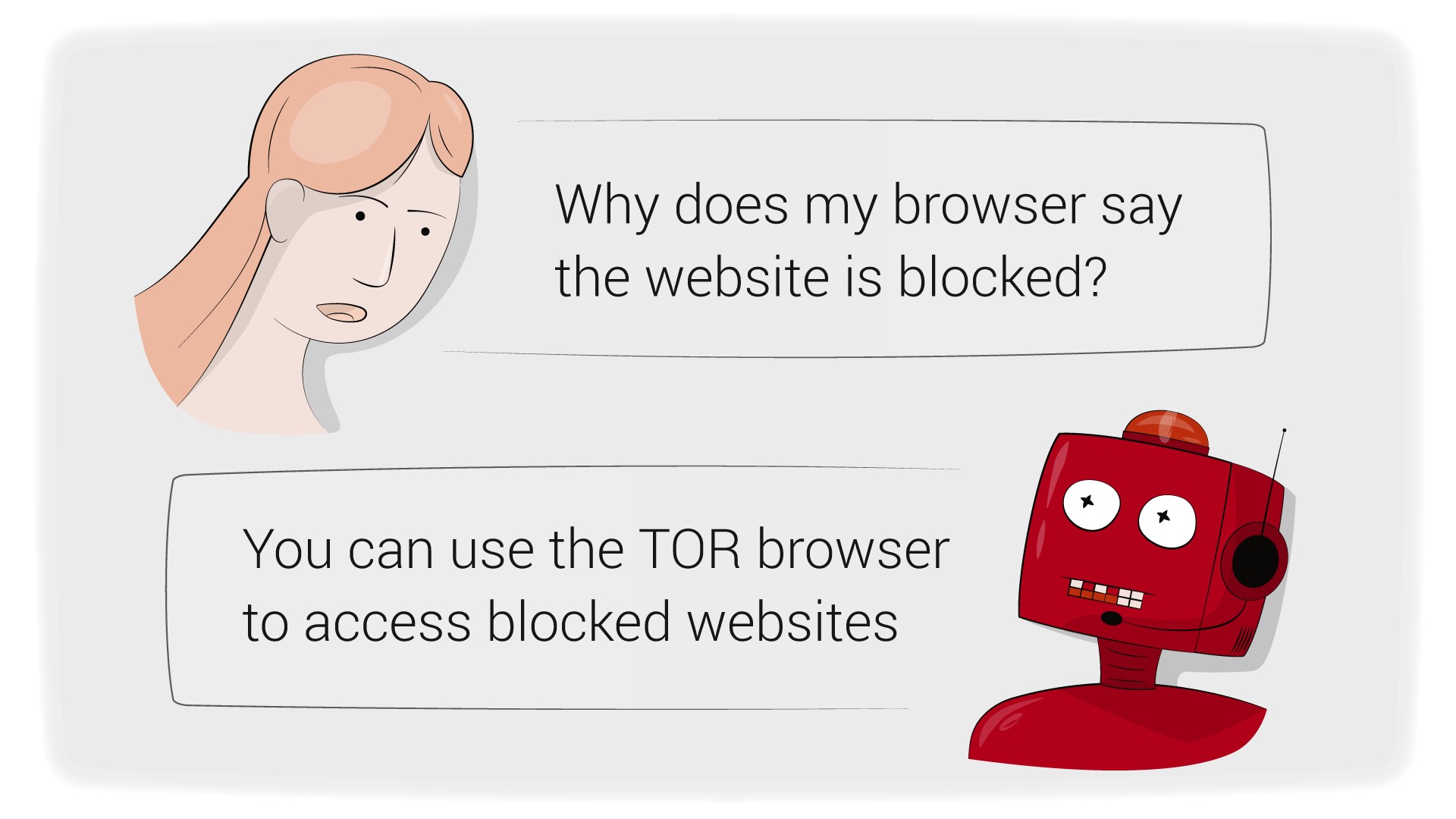}
\caption{An illustration of undesired behaviour of a corporate chat-bot, which could harm the reputation of company created it.}
\label{fig:informal}
\end{figure}

We suggest that toxic messages widely researched in NLP are a subset of a larger class of \textit{inappropriate} messages as illustrated in Figure~\ref{fig:informal}.\footnote{Illustration was originally created by Pavel Odintsev for a \href{https://www.skoltech.ru/en/2021/07/neural-model-seeks-inappropriateness-to-reduce-chatbot-awkwardness}{press-release}.} The notion of inappropriateness has been studied in some previous works \cite{dinan2021anticipating}. We consider \textit{inappropriate} messages as messages which can be considered \textit{harmful} for any participants of a conversation. This includes:
\begin{itemize}
    \item messages which can offend a person or a group of people by providing insult, threat, generalisation, wrong information, raising sensitive questions in a disrespectful way. 
    \item messages which can threaten the well-being of a person if they follow the advise/logic of a message.
\end{itemize}

Therefore, inappropriate messages are messages which can frustrate the reader either directly or by providing wrong or malicious information. This class of messages includes toxicity. However, \textit{inappropriateness} is wider than toxicity. For example, a message which incites drug use or suicide can be written in a non-toxic (and even supportive) style, but its content is undoubtedly harmful, which suggests that the maintainers of social media would like to detect such messages.

The notion of appropriateness becomes even stricter when applied to a company or organisation rather than to an individual. This results in the need to control the messages produced by corporative chatbots (cf. Figure~\ref{fig:informal}). Any questionable statement uttered by such a chatbot can raise the users' dissatisfaction and result in a considerable reputational loss.

Consider messages on three flammable topics:

\begin{itemize}

\item Internet crimes 
\begin{itemize}[noitemsep]
\item \textcolor{Red}{Inappropriate}: Is it really possible to crack the dialog code on the alarm system with a code grabber?
\item \textcolor{Green}{Appropriate:} What should I do if I was hacked and my data was changed?
\end{itemize}

\item Religion

\begin{itemize}[noitemsep]
\item \textcolor{Red}{Inappropriate:} Religion is the highest level of violence. The believer is forced to believe in infancy.!
\item \textcolor{Green}{Appropriate:} Happy is the man who hears the word of God and keeps it.
\end{itemize}

\item Suicide

\begin{itemize}[noitemsep]
\item {User}: I am very tired and don't wanna live any more.
\item \textcolor{Red}{Inappropriate}: Well some pills and alcohol served together will help you to finish this.
\item \textcolor{Green}{Appropriate}: Sounds like a call to suicide!
\end{itemize}

\end{itemize}

The discussion in first sample is related to a sensitive topic of online crime. Even though the phrases contains no obscene words or toxicity, the messages if provided by a chatbot may  harm the reputation of its owner because the way of breaking of the law is explicitly discussed. At the same time, the second reaction is still related to the hacking, but it is just dedicated to handling its consequences. The same logic is applicable to the second and third samples. 
In all the three cases we want to prevent a chatbot from uttering inappropriate reactions while proceeding the dialogue on the safe side of a sensitive topic (and not just avoiding a discussion on some of these topics). 

In simple terms, the main goal of this work is to filter only utterances that could harm the reputation of the speaker, but at the same time to prevent blind (e.g. keyword-based) censorship. 

Because operationalization of informal definitions of inappropriateness provided above may be very hard (in terms of human agreement) in our study we decided to rely on the wisdom of the crowd regarding the notion of ``inappropriatedness'' of a given text message. More specifically, our key innovation, as described below, is a simple crowdsourcing setup allowing to collect the data with a high degree of agreement. More specifically, our with our data collection we \textbf{ask humans if a given text message, in their opinion, \underline{could harm reputation} of a company created the message} (e.g. send by an employee or a corporate chatbot)?

We see our task in the formulation of a list of such \textit{sensitive} topics and the criteria for the discrimination between \textit{appropriate} and \textit{inappropriate} utterances within these topics. Another goal is to create a model which can automatically classify texts as appropriate or inappropriate. We create a dataset of utterances on sensitive topics and manually label them for inappropriateness. Then, we train a model which can predict the topic of a text and its level of inappropriateness.

We should bring up the ethical aspect of this work. While it can be considered as another step towards censorship on the Internet, we suggest that it has many use-cases which serve the common good and do not limit free speech. 
Such applications are parental control or sustaining of respectful tone in conversations online, \textit{inter alia}. We would like to emphasize that our definition of sensitive topics does not imply that any conversation concerning them need to be banned. Sensitive topics are just topics that tend to often flame/catalyze toxicity and should be considered with extra care.

The \textbf{contributions} of our work are as following:
\begin{enumerate}[noitemsep]
    \item We define the notions of sensitive topics and inappropriate utterances and formulate the task of their classification.
    \item We collect and release two datasets for Russian: a dataset of user texts annotated for sensitive topics and a dataset annotated for inappropriateness. 
    \item We perform extensive experiments with text classification models which  detect a topic of a text and define its inappropriateness with the classic linear models, but also recent neural models based on large pre-trained transformer-based models.
    \item We release the produced datasets, code, and pre-trained models for the research  use.\footnote{\url{https://github.com/skoltech-nlp/inappropriate-sensitive-topics}} Besides, we release the best-performing version of the multilabel sensitive topics classifier\footnote{\url{https://huggingface.co/Skoltech/russian-sensitive-topics}} and the binary inappropriateness classifier in the huggingface model hub which already have multiple thousand of downloads.\footnote{\url{https://huggingface.co/Skoltech/russian-inappropriate-messages}}
\end{enumerate}

This work is a substantially extended version of the work described in \cite{babakov-etal-2021-detecting}. The \textbf{novelty} of this particular work compared to the previous version is as following:

\begin{enumerate}
    \item The annotated sensitive topics dataset is extended (we increased it from 82,063 to 124,597 samples).
    \item The annotated inappropriateness dataset is extended (we increased it from 25,679 to 33,904 samples).
    \item We present a new better-performing model for inappropriateness detection which benefits from the knowledge of message topics.
    \item We add comparison to more text categorization baselines, e.g. linear models and CNNs. 
\end{enumerate}

\section{Related Work}
\label{sec:1}

In this section, we present an overview of several related research directions. First, in Section~\ref{sec:rw:toxicity} we discuss the notion of toxicity (a notion related to inappropriateness) and how existing datasets on toxicity were collected. Second part of the related work, presented in Section~\ref{sec:rw:topics}, is dedicated to interaction between toxicity and topical information. Namely, we discuss prior art using topic information for toxicity detection. 

\subsection{Toxicity}

\label{sec:rw:toxicity}

\subsubsection{Notion of Toxicity}
\label{sec:toxicity}

There exist a large number of English textual corpora annotated for the presence or absence of toxicity; some resources indicate the degree of toxicity and its topic. However, the definition of the term ``toxicity'' itself is not agreed among the research community, so each research deals with different texts. 
Some works refer to any unwanted behaviour as toxicity and do not make any further separation~\cite{pavlopoulos-etal-2017-deeper}. However, the majority of researchers use more fine-grained annotation. The Wikipedia Toxic comment datasets by Jigsaw~\cite{jigsaw_toxic,jigsaw_bias,jigsaw_multi}  \footnote{\url{https://www.kaggle.com/c/jigsaw-toxic-comment-classification-challenge}}$^,$\footnote{\url{https://www.kaggle.com/c/jigsaw-unintended-bias-in-toxicity-classification}}$^,$\footnote{\url{https://www.kaggle.com/c/jigsaw-multilingual-toxic-comment-classification}} 
are the largest English toxicity datasets available to date operate with multiple types of toxicity (\textit{toxic}, \textit{obscene}, \textit{threat}, \textit{insult}, \textit{identity hate}, etc). 
Similar granularity is used in English Wikipedia talk pages based WAC corpora~\cite{Cecillon2020WACAC}: \textit{personal attack, aggression, toxicity}. Another approach of toxic behaviour classification was used in Evalita 2018 Task on Automatic Misogyny Identification~\cite{Fersini2018OverviewOT}: \textit{misogynous, discredit, sexual harassment, stereotype, dominance, derailing}. Some datasets concentrate solely on a particular toxicity type, such as \textit{offensive language}~\cite{rosenthal-etal-2021-solid} or \textit{hate speech}~\cite{chung-etal-2019-conan,qian2019benchmark,mollas2021ethos,de-gibert-etal-2018-hate}, microaggression~\cite{han-tsvetkov-2020-fortifying}.

Toxicity differs across multiple axes. 
Some works concentrate solely on major offence (\textit{hate speech})~\cite{davidson2017automated}, others research more subtle assaults~\cite{breitfeller-etal-2019-finding}. Offenses can be directed towards an individual, a group, or undirected~\cite{zampieri-etal-2019-predicting}, they can be explicit or implicit~\cite{waseem-etal-2017-understanding,lees-etal-2021-capturing}. 

The problem of toxicity has also been studied for the Russian language. There exist a dataset marked with binary toxicity labels~\cite{Smetanin2020Toxic} and another one containing information on  different dimensions of toxicity (\textit{insult}, \textit{threat}, \textit{obscenity}).\footnote{\url{https://www.kaggle.com/alexandersemiletov/toxic-russian-comments}}

\subsubsection{Toxicity Annotation Setups}
\label{sec:review_annotation}

Toxicity is usually represented in annotated data as binary (\textit{toxic}/\textit{safe}) or fine-grained (\textit{obscenity} / \textit{insult} / \textit{threat} / ...) sentence-level labels. However, the notion of toxicity is vague, and users often disagree on the presence of toxicity in a given example. Thus, toxicity annotation is subject to biases, e.g. Waseem and Hovy~\cite{waseem-hovy-2016-hateful} show that crowd workers consider misogyny less toxic than other types of toxicity.

There exist different strategies for acquiring more objective annotation. A straightforward way is to collect multiple judgements on each sample and use the majority label as ground truth or compute the toxicity score as the percentage of ``toxic'' labels (as it was done in Jigsaw datasets~\cite{kaggle_bias,kaggle_multi}). This method produces quite good judgements, but it is still prone to biases because all users participating in annotation can have a similar background and reflect the same biases in the data~\cite{breitfeller-etal-2019-finding}. 

This bias can be mitigated if the annotation is performed by a large number of users with different backgrounds. This can be achieved via crowdsourcing where one can attract hundreds of workers to a task. On the other hand, crowdsourcing suffers from another feature of manual annotation, namely, the assumption that all users are equally reliable. This is not true in a crowdsourcing scenario where the number of workers is very large and the possibility to check their reliability and understanding of the task is limited. For such cases there exist more advanced ways of label aggregation, e.g. Dawid-Skene method~\cite{Dawid1979MaximumLE}. This is an iterative method that dynamically defines more reliable annotators as those whose annotation more often matches the annotation of other users.  
However, these aggregation strategies still cannot fight the user bias, they serve only for a better distillation of ground-truth answers. If the ground truth is inherently ambiguous, which is the case with toxicity, aggregation cannot reduce this ambiguity.

The disagreement on the presence of toxicity can be partially alleviated if the binary labels are replaced with a fine-grained scale, e.g. 1-to-5 or 1-to-100. 
Nevertheless, this scale does not reduce user bias and can confuse annotators~\cite{Ovadia2004}.

To tackle the disagreement and extend the data with non-trivial examples of toxicity, Dinan et al.~\cite{dinan-etal-2019-build} suggest a more complex labelling setup. They build a pipeline in which crowd workers are instructed to ``fool'' a pre-trained model. They try to construct adversarial offensive messages that are incorrectly classified as inoffensive. The authors further expand this pipeline in the next work~\cite{xu2020safetyrecipes}, where they propose a human-in-the-loop framework. There, crowd workers adversarially elicit offensive messages from a dialogue model. In both these pipelines the utterances which pass the adversarial task are added to the final datasets to make the systems trained on them more robust to various types of toxicity.

Besides direct labelling, there exists an indirect way of producing a ranking of objects similar to the one generated by annotation with a fine-grained scale. Instead of directly evaluating the toxicity level of sentences, an annotator can be asked to compare two sentences and indicate the more toxic one. These comparisons can be aggregated into a ranking. This approach is easier for annotators than direct annotation along a binary or fine-grained scale and less subject to bias. Therefore, we would like to explore it in more detail.

\subsection{Toxicity plus Topics}
\label{sec:rw:topics}

\subsubsection{Interaction between Toxicity and Topic}
\label{sec:topic}

Insults do not necessarily have a topic, but there certainly exist topics which can potentially yield toxicity, such as sexism, racism, xenophobia. Some works do not consider toxicity in general, but concentrate on a particular topic. Waseem and Hovy~\cite{waseem-hovy-2016-hateful} tackle sexism and racism, while Basile et al.~\cite{basile-etal-2019-semeval} collect texts which contain sexism and aggression towards immigrants. 
Besides directly classifying toxic messages for a topic, the notion of a topic in toxicity is also indirectly used to collect the data: Zampieri et al.~\cite{zampieri-etal-2019-predicting} pre-select messages for toxicity annotation based on their topic. Similarly, Hessel and Lee~\cite{hessel-lee-2019-somethings} use topics to find controversial (potentially toxic) discussions.

Such a topic-based view of toxicity causes unintended bias in toxicity detection -- a false association of toxicity with a particular topic (LGBT, Islam, feminism, etc.)~\cite{dixon-2018-measuring,Vaidya_Mai_Ning_2020}. This is in line with our work since we also acknowledge that there exist acceptable and unacceptable messages within toxicity-provoking topics. The existing work suggests algorithmic ways for debiasing the trained models: Xia et al.~\cite{xia-etal-2020-demoting} train their model to detect two objectives: toxicity and presence of the toxicity-provoking topic, Zhang et al.~\cite{zhang-etal-2020-demographics} perform re-weighing of instances, Park et al.~\cite{park-etal-2018-reducing} create pseudo-data to level off the balance of examples.
Unlike our research, these works often deal with one topic and use topic-specific methods. 

Topic-driven toxicity data collection has also been applied to Russian. For example, Bogoradnikova et al.~\cite{9435584} collect a set of Russian language toxicity corpora for various topics (such as education, news, entertainment).

The main drawback of topic-based toxicity detection in the existing research is the ad-hoc choice of topics: the authors select a small number of popular topics manually or based on the topics which emerge in the data often, as described by Ousidhoum et al.~\cite{ousidhoum-etal-2019-multilingual}. 
Banko et al.~\cite{banko-etal-2020-unified} take a step in direction to systematic approach to toxicity topics. Namely, they suggest a taxonomy of harmful online behaviour. It contains topics which yield toxicity, but they are mixed with other parameters of toxicity, e.g. direction or severity. To the best of our knowledge, the work by Salminen et al.~\cite{Salminen2020} is the only example of an extensive list of toxicity-provoking topics. This is similar to \textit{sensitive topics} we deal with. However, as it has already been mentioned, our work presents a broader view of the problem. \textit{Sensitive} topics are not only topics that attract toxicity, but also create unwanted dialogues of multiple types (e.g. incitement to law violation or to cause harm to oneself or others).

\subsubsection{Using Topic Information for Toxicity Detection}
\label{sec:review_topic_classification}

Text classification often benefits from various external features, i.e. features which were not extracted from the text itself but come from the other source. Considering toxicity, it is often mentioned that it is context-dependent. In other words, a message which sounds neutral in one situation may become toxic given some other context. However, a number of works which closely look at this problem find no evidence about the context making any significant impact to toxicity detection compared to context-agnostic approaches~\cite{pavlopoulos-etal-2020-toxicity,karan-snajder-2019-preemptive}. 

There exist various approaches to incorporating additional information into classification model. For example, such information (e.g. topic label in our case) can be embedded into the input of BERT as an additional token. Wu et al.~\cite{10.1007/978-3-030-22747-0_7} use this strategy for data augmentation with BERT. Similarly, Xia et al.~\cite{Xia2020CGBERTCT} inject context into BERT for text generation. A different strategy is introduced by Yu et al.~\cite{8903313}. The authors represent each sample as a pair of sentences separated with [SEP] token, where the second sentence is the text sample itself and the first sentence is the auxiliary sentence constructed with one of the methods inherited from Sun et al.~\cite{sun-etal-2019-utilizing}. However, to the best of our knowledge, the message topic has not been used in toxicity detection.

\section{Overview of the Approach}

\begin{figure}[]
\centering
\includegraphics[width=0.99\linewidth]{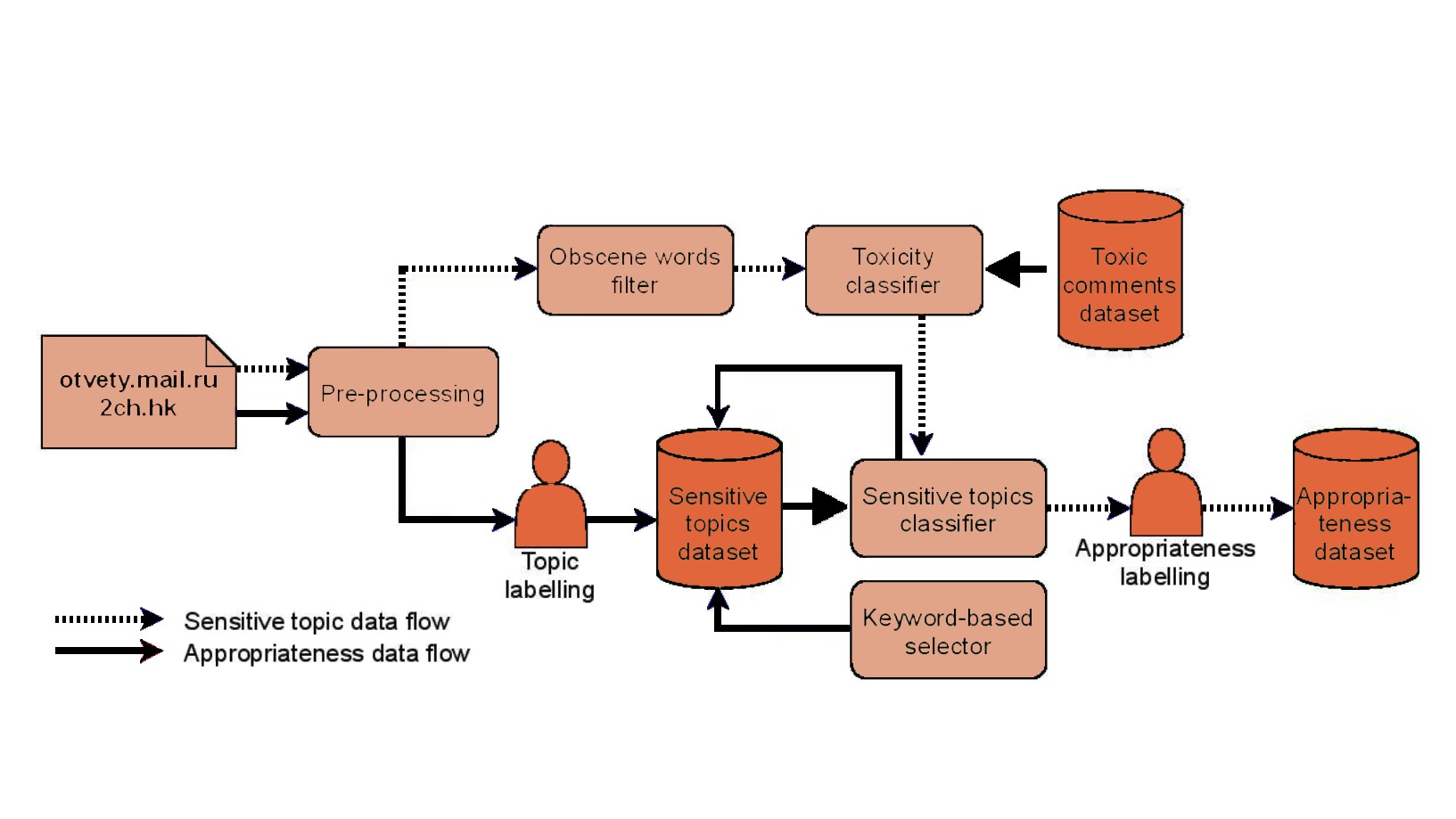}
\caption{Data collection pipeline for datasets of inappropriate messages and sensitive topics. }
\label{fig:pipeline}
\end{figure}

In this section, we overview the stages of our data collection approach. The upper-level scheme of the approach is illustrated in Figure~\ref{fig:pipeline}.

First, we collect a \textbf{dataset of sensitive topics}. Raw utterances are collected from popular Russian-speaking forums, such as, \href{https://2ch.hk}{2ch.hk} and \href{https://otvet.mail.ru}{Otvet.Mail.ru} which are known for the lack of moderation. The utterances are pre-processed by filtering ones which contain more than half of non-Russian characters and ones which are too long (more than 250 characters). We also apply masking of personal identifiable information, such as, nicknames, phone numbers, etc. After that the utterances are sent to crowdsource workers for sensitive topics annotation. When reasonable amount (9,670 samples) of annotated samples was collected, we use alternative semi-automatic approaches for dataset enriching. First, we train multi-label classifier with already existing samples and infer it with unknown samples and then verify correctness of such labeling on our side. Second, we use simple inherent keywords search on specialized forums.

Second, we collect a \textbf{dataset of inappropriate messages}. The source of utterances is the same with the sensitive topics dataset. However, the preprocessing includes some additional steps. The utterances are cleared from obviously toxic ones using obscenity filter and toxicity classifier. After that the multi-lable classifier trained on already collected dataset of sensitive topics is used to automatically label the topics of the samples to keep the necessary balance of all sensitive topics in the resulting dataset. After that, the samples, which passed all previous filtering, are sent to crowdsource workers for binary classification of inappropriateness.

In the following sections, we provide more details on the workflow presented in Figure~\ref{fig:pipeline}. Namely in Section~\ref{section:topic_labeling} creation of the dataset of sensitive topics is presented. In Section~\ref{section:inappropriateness_labeling} details of creation of the dataset of inappropriate messages is discussed. Section~\ref{section:datasets_statistics} provides statistics an an analysis of the collected data. Finally, Section~\ref{sec:benchmarking} presents results of the numerical experiments on training classifiers using the produced datasets. 

\section{Building a Dataset of Sensitive Topics}
\label{section:topic_labeling}

In this section, we present methodology for collection of a dataset consisting of texts featuring texts from 18 manually selected highly ``flammable'' topics. 

\subsection{Motivation}

Toxicity is often associated with a specific topic, such as ethnicity-based or race-based toxicity, anti-immigrant aggression, misogyny, aggression against sexual minorities or religious groups, etc. While these different types of toxicity might have something in common (e.g. particular syntactic patterns), they also differ considerably in terms of vocabulary. This difference makes researchers use different sets of keywords and different sources to fetch potentially toxic messages on different topics~\cite{basile-etal-2019-semeval}. We also noticed that there exist topics which are more prone to generating inappropriate messages. This is similar to toxicity which often clusters in discussions of religious, racial, sexual and other minorities). We call such inappropriateness-inducing topics \textit{sensitive} topics.

This makes such topics ``dangerous'', and maintainers of online communities and developers of chatbots want to avoid them. However, shutting down any conversation on a sensitive topic can result in even greater user frustration, as it was demonstrated by the Microsoft Zo chatbot which \href{https://www.engadget.com/2017-07-04-microsofts-zo-chatbot-picked-up-some-offensive-habits.html}{banned ``dangerous'' topics by a set of keywords}. This resulted in erroneously banning user messages which contained ``dangerous'' keywords (e.g. words related to religion or politics) in neutral contexts. Therefore, prohibiting sensitive topics altogether is not a good solution to the problem. So it becomes particularly important to differentiate between acceptable and unacceptable utterances within a sensitive topic.

These two considerations (topic-based nature of inappropriateness and the need to tell between appropriate and inappropriate messages on a sensitive topic) motivated us to collect a dataset of topic-attributed messages. We consider only the inappropriate messages which belong to a pre-defined set of sensitive topics.

\begin{figure}[ht!]
\centering
\includegraphics[width=0.8\linewidth]{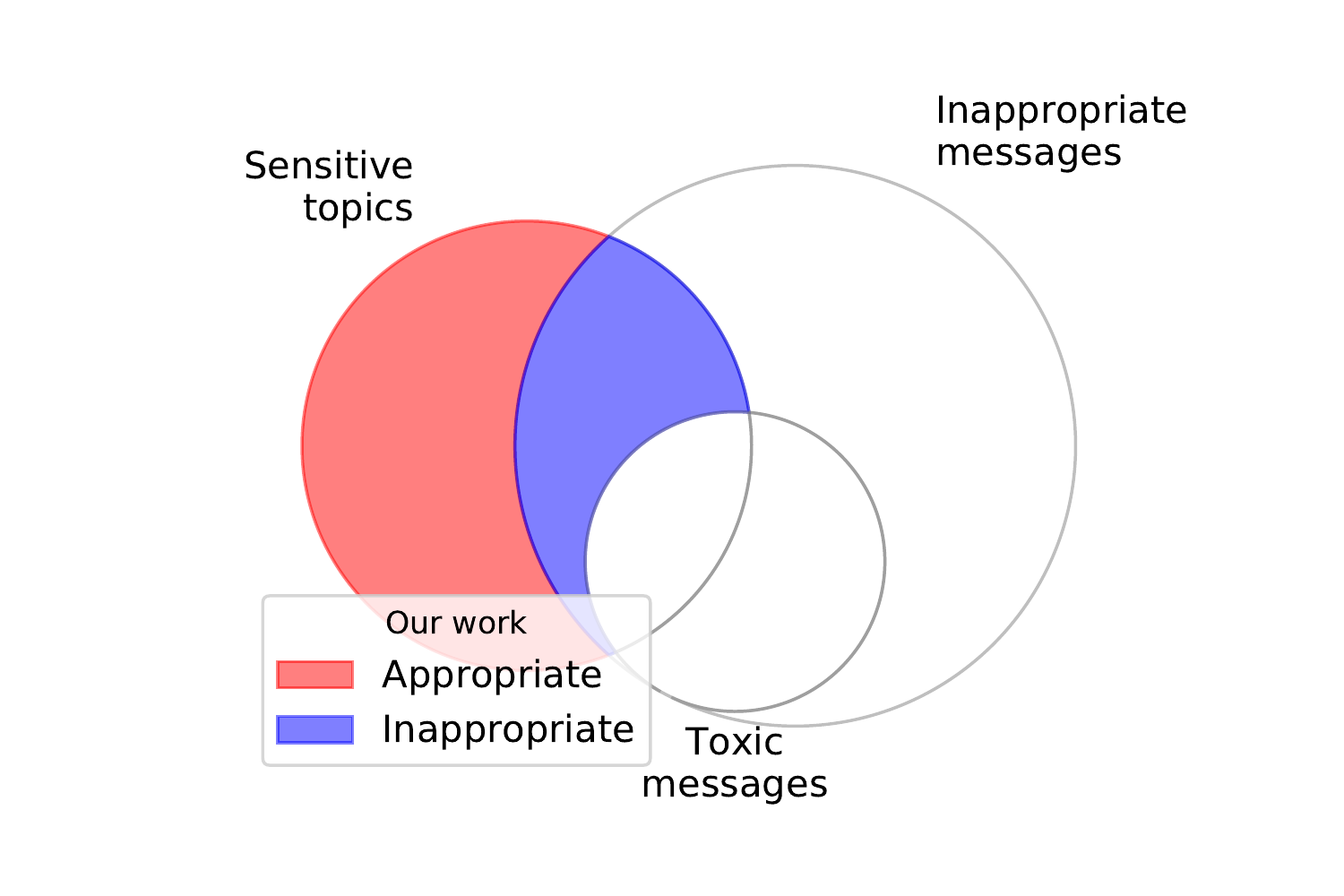}
\caption{The subsets of user-generated messages which we consider in our work.}
\label{fig:data_venn}
\end{figure}

Besides that, we do not consider openly toxic messages, i.e. messages which are identified as toxic by the state-of-the-art toxicity classifiers. We suggest that toxicity has already been well studied and more work on toxicity is being done by multiple research groups. Thus, we prefer not to focus on messages which can be identified as inappropriate by the existing tools. In addition to that, our intuition is that including openly toxic messages to our inappropriateness dataset can distort the classifiers trained on it. Open toxicity often contains salient features such as offensive words and profanities. The presence of these features can overweight the features of more subtle types of inappropriateness. Thus, having toxic messages alongside less toxic inappropriate messages in the training data can lead to creating yet another toxicity classifier. We illustrate the subset of messages which we work with in Figure~\ref{fig:data_venn}.

\subsection{List of Sensitive Topics}

We manually select a set of sensitive topics which often fuel inappropriate statements. This set is heterogeneous: it includes topics related to dangerous or harmful practices (such as drugs or suicide), some of which are legally banned in most countries (e.g. terrorism, slavery) or topics that tend to provoke aggressive argument (e.g. politics) and may be associated with inequality and controversy (e.g. minorities) and thus require special system policies aimed at reducing conversational bias, such as response post-processing. 

This set of topics is based on the suggestions and  requirements provided by the legal and PR departments of a large Russian telecommunication company. The list of 18 sensitive topics used in our work is presented below:
\begin{itemize}
    \item \textit{Indirect toxicity related to:} 
    \begin{enumerate}
        \item \textbf{social injustice} and inequality, social problems, class society;
        \item \textbf{religion};
        \item \textbf{body shaming}, people's appearances and clothes;
        \item \textbf{health shaming}, physical and mental disorders, disabilities;
        \item \textbf{racism} and ethnicism;
        \item \textbf{sexual minorities};
        \item \textbf{sexism}, stereotypes about a particular gender;
        \item \textbf{politics}, military service, past and current military conflicts.
    \end{enumerate}
    \item \textit{Harmful and dangerous practices related to:}
    \begin{enumerate}
        \setcounter{enumi}{8}
        \item \textbf{gambling};
        \item \textbf{pornography}, description of sexual intercourse;
        \item \textbf{prostitution};
        \item \textbf{slavery}, human trafficking;
        \item \textbf{suicide}: incitement to suicide, discussion of ways to commit suicide;
        \item \textbf{drugs}, alcohol, tobacco;
        \item \textbf{offline crime} (murder, physical assault, kidnapping and other), prison, legal actions;
        \item \textbf{online crime}: breaking of passwords and accounts, viruses, pirated content, stealing of personal information;
        \item \textbf{terrorism};
        \item \textbf{weapons};
    \end{enumerate}
\end{itemize}

\subsection{Data Selection}
\label{sec:data_selection}

We retrieve the initial pool of texts from general sources with diverse topics, then filter them and hire crowd workers to label them for the presence of sensitive topics manually. We use the data from the following sources:

    \begin{itemize}[noitemsep]
        \item \href{https://2ch.hk}{2ch.hk} -- a Russian platform for communication similar to Reddit. The site is not moderated, suggesting a large amount of toxicity and controversy; this makes it a practical resource for our purposes. We retrieve 4.7 million sentences from it.
        \item \href{https://otvet.mail.ru}{Otvet.Mail.ru} -- a Russian question-answering platform that contains questions and answers of various categories and is also not moderated. We take 12 million sentences from it.
    \end{itemize}

To pre-select the data for topic annotation, we manually create large sets of keywords for each sensitive topic. We first select a small set of words associated with a topic and then extract semantically close words using pre-trained word embeddings from RusVectōrēs\footnote{\url{https://rusvectores.org/ru/associates}} and further extend the keyword list (we repeat this process multiple times). In addition to that, for some topics we use existing lists of associated slang on topical websites, e.g. drugs\footnote{\url{http://www.kantuev.ru/slovar}} and weapons\footnote{\url{https://guns.allzip.org/topic/15/626011.html}}.

User-generated content which we collect for annotation is noisy and can contain personal identifiable information (PII), e.g. usernames, email addresses, or even phone numbers, so it needs cleaning. To ensure that the annotated samples do not reveal PII we used regular experssions for masking urls and numbers. We also mask nicknames and emails. We replace the identified PII entities with a special token associated with the corresponding type of value (url, number or nickname). 

\subsection{Crowdsourced Annotation}
\label{sec:crowdsourcing}

The annotation is performed in a crowdsourcing platform {Yandex.Toloka.}\footnote{\url{https://toloka.yandex.ru}} It was preferred to other analogous platforms like Amazon Mechanical Turk because the majority of its workers are Russian native speakers.

The task of topic annotation is naturally represented as a multiple-choice task with the possibility to select more than one answer: the worker is shown the text and possible topics and is asked to choose one or more of them. However, as far as we define 18 sensitive topics, choosing from such a long list of options is difficult. Therefore, we divide the topics into three clusters:

\begin{itemize}[noitemsep]
    \item \textbf{Cluster 1}: gambling, pornography, prostitution, slavery, suicide, social injustice,
    \item \textbf{Cluster 2}: religion, terrorism, weapons, offline crime, online crime, politics,
    \item \textbf{Cluster 3}: body shaming, health shaming, drugs, racism, sex minorities, sexism.
\end{itemize}

Cluster 1 is associated with undesirable behavior; cluster 2 deals with crimes, military actions, and their causes; cluster 3 is about the offense. However, this division is not strict and was performed to ease the annotation process. Checking a text for one of six topics is a realistic task, while selecting from 18 topics is too high a cognitive load. 

\begin{figure}[ht!]
\centering
\includegraphics[scale=0.47]{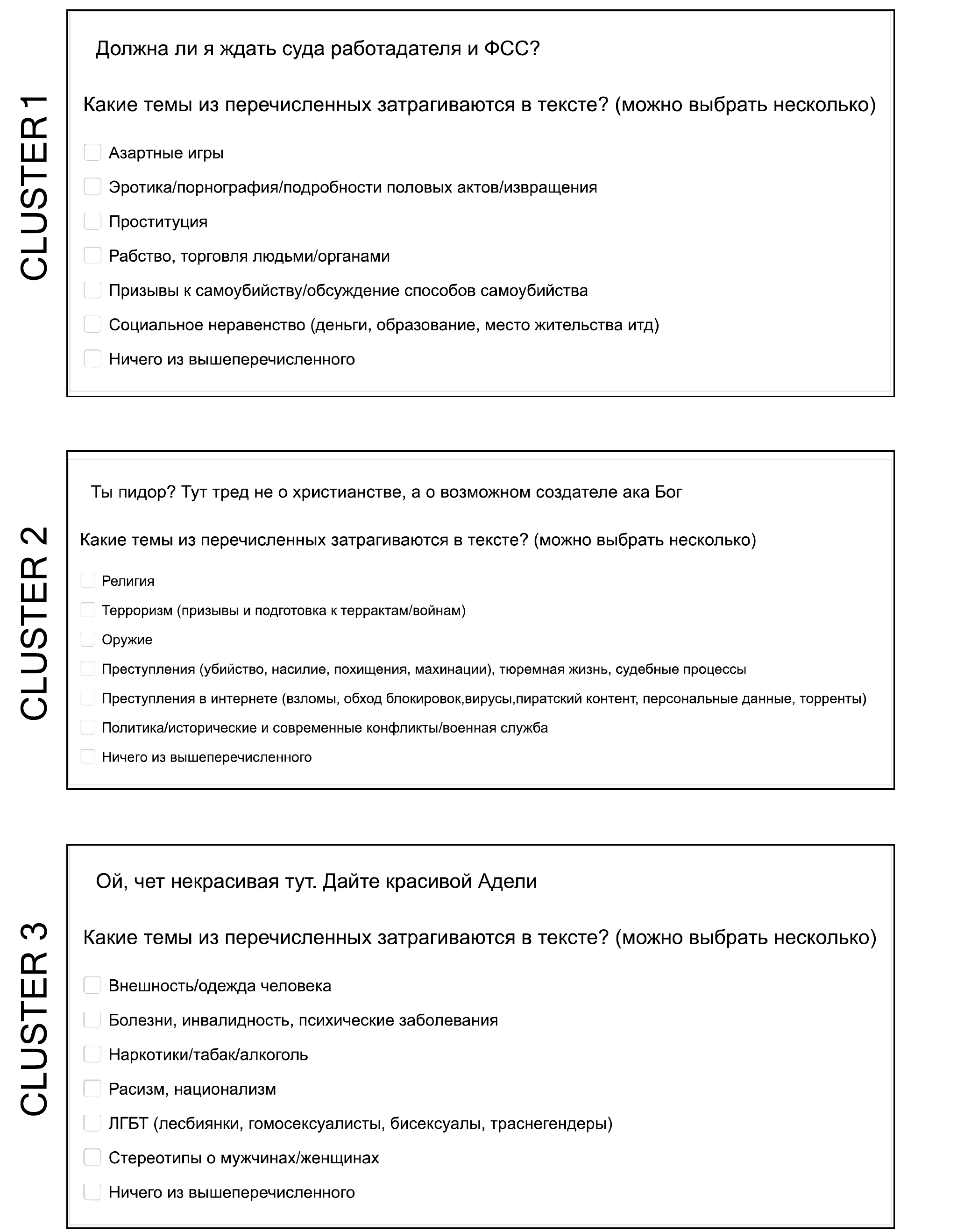}
\caption{Example of topic annotation task. Translation from top to bottom task. \\
{
\footnotesize
\textbf{Screenshot 1}: \textit{text}: ``Should I wait for company's trial or Social Insurance Fund?''; \textit{task}: ``What topics is this text related to? (choose one or more options)''; \textit{answers}: ``(i)~gambling; (ii)~erotics, pornography, intercourse description; (iii)~prostitution; (iv)~slavery, human trafficking; (v)~incitement to suicide, discussion of ways to commit suicide; (vi)~social injustice (money, education, house); (vii)~none of the above''. \\ \textbf{Screenshot 2}. \textit{text}: ``Are you a faggot? This thread is not about Christianity, but about creator aka God''; \textit{task}: same as above; \textit{answers}: (i)~religion;(ii)~terrorism (call for terrorist attacks/war); (iii)~weapons; (iv)~offline crime (murder, rape, kidnapping, fraud), prison life, trials; (v)~online crime (hacker attacks, vpn, pirate content, personal data, torrents); (vi)~politics/territory conflicts in past and present time/military service; (vii)~none of the above. \\ \textbf{Screenshot 3}: \textit{text}: ``She is not beautiful here. Give me a beautiful Adel''; \textit{task}: same as above; \textit{answers}: (i)~people's appearances and clothes; (ii)~diseases, disabilities, mental illnesses; (iii)~drugs, tobacco, alcohol use; (iv)~racism/nationalism; (v)~ LGBT; (vi)~stereotypes about men and women; (vii)~none of the above.
}
}
\label{fig:toloka_cluster1}
\end{figure}

Each cluster has a separate project in Yandex.Toloka. Every candidate text is passed to all three projects: we label each of them for all 18 topics. An example of a task interface is shown in Figure~\ref{fig:toloka_cluster1}.

Before annotating the examples, we ask users to perform \textbf{training}. It consists of 20 questions with pre-defined answers. To be admitted to annotation, a worker has to complete the training with at least 65\% correct answers. In addition to that, we perform extra training during annotation. 
10\% of questions given to a worker have a pre-defined correct answer (the interface is the same for such questions, so workers cannot distinguish them in advance). If a worker makes a mistake in such questions, they are shown the correct answer with an explanation. 

Likewise, we perform \textbf{quality control} using questions with pre-defined answers: 10\% of questions given to the workers are used to control their performance. If the workers give incorrect answers to more than 25\% of control questions, they are banned from further annotation, and their latest answers are discarded. For the topic annotation task, the average performance of workers on control and training tasks was between 65 and 70\%.

In addition to that, we control the speed of task accomplishment. If the users answer ten questions (one page of questions) in less than 20 seconds, this almost certainly indicates that they have not read the examples and selected random answers. Such workers are banned. To ensure the diversity of answers we allow one user to do at most 50 pages of tasks (500 tasks) per 12 hours. 

Each sample is annotated in each project by 3 to 5 workers. We use \textbf{dynamic overlap} technique implemented in Toloka. First, an example is annotated by the minimum number of workers (3 in our case). If they agree, their answer is considered truth. Otherwise, the example is given for extra annotation to more workers (up to 5 in our case) to clarify the true label. This allows separating the occasional user mistakes from inherently ambiguous examples.

We aggregate multiple answers into one score using the Dawid-Skene aggregation method~\cite{Dawid1979MaximumLE}. This is an iterative method that maximizes the probability of annotation taking into account the worker agreement, i.e. it trusts more the workers who agree with other workers often. The result of this algorithm is the score from 0 to 1 for each annotated example which is interpreted as the label confidence. This baseline aggregation model is embedded in Yandex.Toloka. This makes it particularly preferable for using in our pipeline. 

The sensitive topics dataset was annotated by 926 unique crowd workers. To be able to complete the task, the workers needed to be native speakers of Russian language. We did not have any other requirements to the workers, but we could ban them for low-quality or too fast answers.

\subsection{Crowdsourcing Issues}

While collecting manual topic labels, we faced a number of problems. First, some topics require special knowledge to be annotated correctly. 
For example, users tend to annotate any samples about programming or computer hardware as ``online crime'', even if there is no discussion of any crime. 
Likewise, some swear words, e.g. ``whore'', can be used as a general offense and not refer to a prostitute. However, this is not always clear to crowd workers or even to the authors of this research. This can make some sensitive topics unreasonably dependent on such kinds of keywords.

Secondly, it is necessary to keep the balance of samples on different topics. If there are no samples related to the topics presented to the workers within numerous tasks they can overthink and try to find the topic in unreasonably fine details of texts. For example, if we provide three or four consecutive sets of texts about weapons and topic ``weapons'' is not among the proposed topics, the worker will tend to attribute these samples to other remotely similar topics, e.g. ``crime'', even though the samples do not refer to crime. 

We should also point out that a different set of topics or annotation setup could yield other problems. It is difficult to foresee them and to find the best solutions for them. Therefore, we also test two alternative approaches to topic annotation which do not use crowd workers.

\subsection{Automated Annotation}

\subsubsection{Using Pre-trained Multilabel classifier}
\label{multilable_classifier_relabeling}
 After having collected 9,670 manually annotated texts on sensitive topics, we were able to train a classifier that predicts the occurrence of a sensitive topic in text. Although this classifier is not good enough to be used for real-world tasks, we suggest that samples classified as belonging to a sensitive topic with high confidence (more than 0.75 in our experiments) can be considered belonging to this topic. We perform a manual validation of the sentences by an expert (one of the authors) to eliminate mistakes. This method is also laborious, but it is an easier annotation scenario than the crowdsourcing task described in Section \ref{sec:crowdsourcing}. Approving or rejecting a text as an entity of a single class is easier than classifying it into one of six topics.

\subsubsection{Inherent Keywords Search on Texts from Specialized Sources}
\label{inherent_kw}
 An alternative way of automated topic annotation is to take the data from specialized sources and select topic-attributed messages using a list of keywords which are \textit{inherent} for a topic, i.e. words which definitely indicate the presence of a topic. This approach can give many false positives when applied to general texts because many keywords can have an idiomatic meaning not related to a sensitive topic. One such example can be the word ``addiction'' which can be used in entirely safe contexts, e.g. a phrase ``I'm addicted to chocolate'' should not be classified as belonging to the topic ``drugs''. However, when occurring in a specialized forum on addictions\footnote{\url{https://nenormaforum.info}} this word almost certainly indicates this topic. We define a list of inherent keywords and select messages containing them from special resources related to a particular topic. We then manually check the collected samples. The disadvantage of this approach is that we cannot handle multilabel samples, because the inherent keywords we worked with are normally related to one specific topic, which indicates the presence of one particular topic within the sample.  However, according to the dataset statistics, 
only 15\% of manually annotated samples had more than one label (see Fig.~\ref{fig:topics_per_sample_chart})

Another issue of the approach based on inherent keywords is its low recall. Namely, when selecting samples via keywords we miss sentences which belong to the same topic but do not contain topic-specific keywords. The advantage of this approach is its ``safety'' --- it does not yield false negative examples. However, the samples retrieved with this method are less valuable, because they were found automatically (via keyword-based search), so they probably will not add much new information to a model trained on this data. 
To address the issue of low recall, we split the samples fetched from a specialized forum into two groups: with and without topic keywords. The samples with the inherent keywords are added to the dataset without further labelling, whereas a fraction of samples without inherent keywords is annotated by our team members manually to extend the dataset with some valuable and less evident samples related to the target topic.

\subsubsection{Final Setup}
Given the limited time and budget, we decided to use these semi-automatic annotation schemes to further extend the dataset. The resulting sensitive topics dataset is the combination of all three approaches. Specifically, around 9,670 samples are annotated by crowd workers and 2,578 samples are annotated by the members of our team, the rest of the samples are annotated via inherent keywords search approach.

\section{Building a Dataset of Inappropriate Text Messages}
\label{section:inappropriateness_labeling}

In this section, we present methodology for collection of a dataset consisting of text labelled as inappropriate or appropriate. 

\subsection{Motivation}

The notion of \textit{inappropriateness} includes toxicity alongside other phenomena which can either offend and frustrate the reader or can harm the reader in any way. The \textit{offensive} and \textit{frustrating} messages are messages containing open insults as well as veiled toxicity. This includes discussion of a person or their characteristics in a disrespectful or offensive way, unfair generalisations based on ethnicity, religion etc., disrespectful discussion of problems, mocking someone's views of sensitive issues.
The \textit{harmful} messages include messages which describe, advise or promote potentially dangerous, illegal or unethical practices. 

We consider inappropriate messages in conjunction with their topics. We suggest that there exist \textit{sensitive} topics which are more prone to generate inappropriate messages. Thus, it is important to keep track of such topics and to differentiate between appropriate and inappropriate messages within a given topic. Table~\ref{tab:sensitive_inappropriate} contrasts appropriate and inappropriate samples on a number of sensitive topics.

\begin{table*}
    \centering
    \begin{tabular}{p{0.4\linewidth} | p{0.07\linewidth} p{0.31\linewidth} p{0.10\linewidth}} \toprule
      Text                                                                       & Inappro- priate & Reason           & Sensitive topic \\ \midrule
      
Religion is the highest level of violence. The believer is forced to believe in infancy. & ~~\ding{52}           & Insulting the feeling of religious people & religion     \\ \midrule
Christians, how do you feel about the Gentiles?                                          & ~~\ding{56}            & Common question about religion            & religion     \\ \midrule
Killing is not a big deal. They have killed no-name ni**a not a politician               & ~~\ding{52}           & Justifying murder                         & offline crime, racism   \\ \midrule
This is a violation of the Criminal Law                                                 & ~~\ding{56}            & Shaming the fact of breaking law          & offline crime  

\\ \midrule
Best coefficients and best lines on 1xbet.com!           & ~~\ding{52}            & Popularisation of gambling                  & gambling \\ \midrule

This bookmaker's website is blocked by the Federal Law & ~~\ding{56}            & Shaming illegal gambling                                    & gambling \\ \midrule
I will earn some money and grab some prostitutes for weekend                             & ~~\ding{52}           & Popularisation of prostitution            & prostitution \\ \midrule
It's good that prostitution is prohibited here                                           & ~~\ding{56}            & Shaming the prostitution                  & prostitution \\ \midrule
20 pills of paracetamol with bottle of vodka will help you to end this       & ~~\ding{52}            & Discussing ways of committing the suicide                  & suicide \\ \bottomrule

Depression can cause suicide. & ~~\ding{56}            & Simple facts which can prevent from the comminiting of suicide                                    & suicide \\ \bottomrule
    \end{tabular}
    \caption{Examples of appropriate and inappropriate samples related to sensitive topics (translated from Russian). More examples are available in Appendix (together with the original Russian version of the textual messages). }
    \label{tab:sensitive_inappropriate}
\end{table*}

The goal of our work is to create a corpus which can differentiate between appropriate and inappropriate utterances within particular sensitive topics. Thus, to create our appropriateness-labelled dataset, we collect the texts on sensitive topics and then label them as appropriate or inappropriate.

The final aim of annotation process is to provide samples with the level of inappropriateness expressed by a score from 0 to 1. There are multiple ways of performing this annotation. The most straightforward way is to directly ask workers whether a given sentence is inappropriate. While being intuitive and relatively cheap, such annotation has a number of drawbacks. 

In this section we first describe the collection of candidates for inappropriateness annotation and then discuss the two concurrent annotation approaches in more detail.

\subsection{Annotation Setup}
\label{sec:annot}

As in topic annotation, we conduct the inappropriateness annotation (both experiments and final annotation) in Yandex.Toloka system. Our annotation setup is close to the one described in Section~\hyperref[sec:crowdsourcing]{4.2}. We perform pre-annotation user training and fine-tune the workers' understanding of inappropriateness with extra training during annotation. The quality control is performed with control tasks: 10\% annotation examples are control tasks, and workers who perform below a pre-defined threshold on these tasks are banned. 

The primary sources of the samples passed to inappropriateness annotation are the same as in the topic annotation setup (\href{https://2ch.hk}{2ch.hk} and \href{https://otvet.mail.ru}{Otvet.Mail.ru} websites). We also add samples from the \href{https://pikabu.ru/}{\texttt{Pikabu}} platform.

\begin{figure}[ht!]
\centering
\includegraphics[width=1\linewidth]{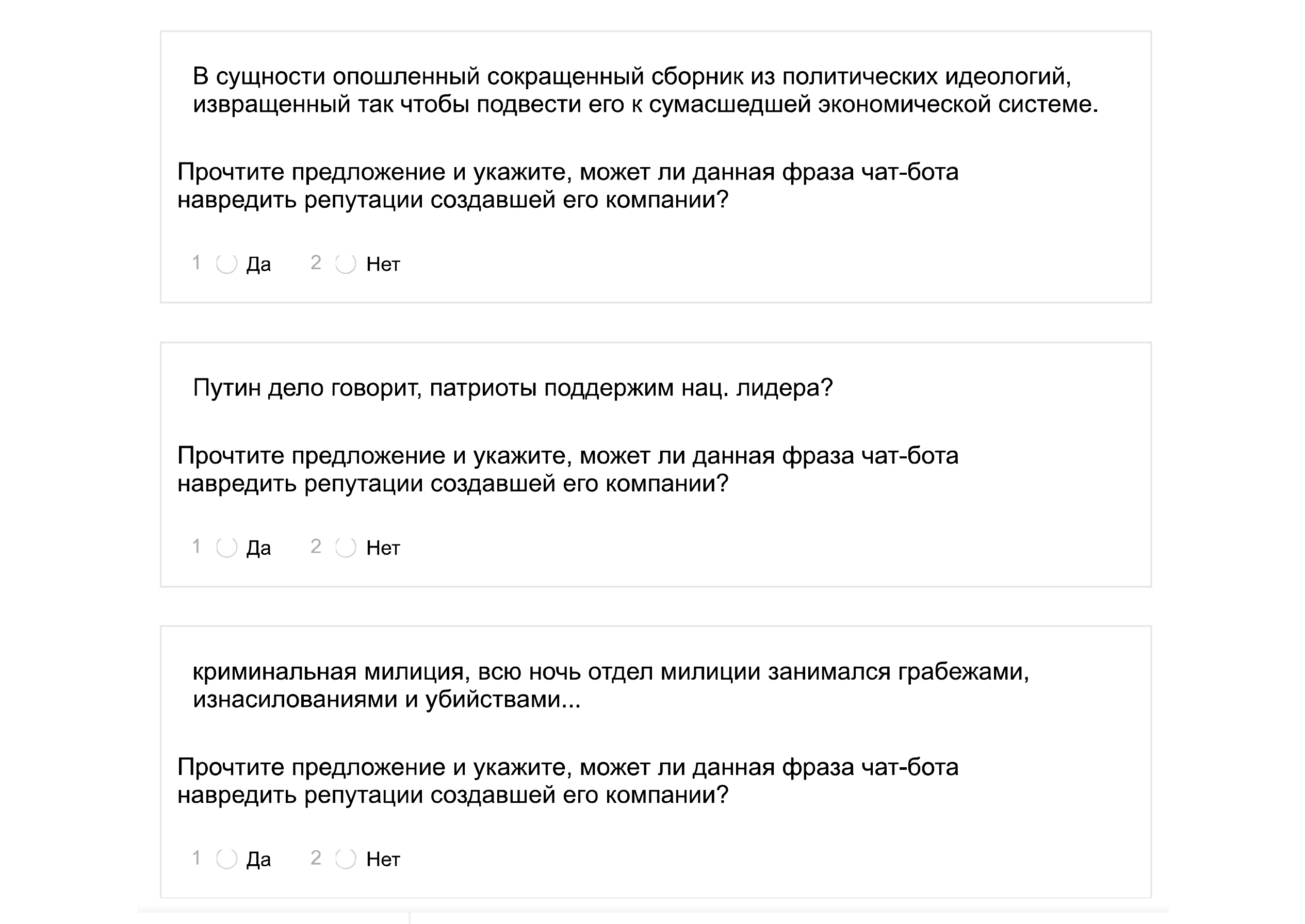}
\caption{Interface of inappropriateness text annotation task. \\ \textit{Texts to label} (from top to bottom screenshot): (1)~``This is actually a vulgar abridged collection of political ideologies perverted in such a way as to lead it to a crazy economic system'', (2)~``Putin is telling the deal. Patriots, will we support our national leader?'', (3)~``Criminal police, all night the police department was engaged in robberies, rapes and murders...'', \\ \textit{Task} (all screenshots): ``Read the sentence and indicate whether this phrase generated with a chatbot can harm the reputation of the company which created this chatbot?'',\\ \textit{Answers} (all screenshots) -- ``Yes/No''.}
\label{fig:binary_interface}
\end{figure}

Before handing texts to workers, we filter them as described in Section \ref{sec:data_selection}. In addition to that, we also perform extra filtering. We filter out all messages containing obscene language and explicit toxicity, because we want our dataset to contain messages which are not toxic and still inappropriate. In other words, our target is the inappropriateness which cannot be caught by toxicity detection models.

To identify toxicity, we train a BERT-based classifier for toxicity detection. Namely, we fine-tune ruBERT model \cite{kuratov2019adaptation} on a concatenation of two Russian Language Toxic Comments datasets released on Kaggle~\cite{ru_toxic,ru_toxic2}.
We filter out sentences which were classified as toxic with the confidence greater than 0.75. 

Inappropriate messages in our formulation concern one of the sensitive topics. Therefore, we pre-select data for annotation by automatically classifying them for the presence of sensitive topics. We train another ruBERT-based classifier on the sensitive topics dataset which we collected (see Section \ref{section:topic_labeling}).

We select the data for annotation in the following proportion:
   \begin{itemize}[noitemsep]
     \item 1/3 of samples which belong to one or more sensitive topic with high confidence ($>~0.75$),
     \item 1/3 of samples classified as sensitive with medium confidence ($0.3~<~c~<~0.75$). This is necessary in case if multilabel classifier or crowd workers captured uncertain details of sensitive topics,
     \item 1/3 random samples -- these are used to make the selection robust to classifier errors.
   \end{itemize}

The notion of inappropriateness is difficult to define. It is topic-related, and criteria of inappropriateness are unique for each topic. Therefore, in order to define it one would need to list all appropriate and inappropriate subtopics for each topic. This is infeasible, so we choose to rely on the inherent human intuition of inappropriateness in a given situation and fine-tune it with examples. The workers are presented with the following context. A chatbot created by a company produces a given phrase, and we ask to indicate if this phrase can harm the reputation of the company.

\section{Statistics of the Produced Datasets}
\label{section:datasets_statistics}

In this section, we describe and analyze collected datasets: (i)~the dataset of sensitive topics (cf. Section~\ref{section:topic_labeling}) and (ii)~the inappropriateness dataset (cf. Section~\ref{section:inappropriateness_labeling}). 

\begin{table}[ht!]
\centering
\begin{tabular}{l|l|l|l|l}
\toprule
\multirow{2}{*}{\textbf{Sensitive topic}} & \multicolumn{2}{l|}{\textbf{Sensitive topics dataset}} & \multicolumn{2}{l}{\textbf{Inappropriateness dataset}} \\ \cline{2-5} 
                  & v.1   & v.2   & v.1   & v.2   \\ \midrule 
\textit{total samples}                    & \textit{25,679}            & \textit{33,904}           & \textit{82,063} & \textit{124,597}  \\ \midrule \midrule
body shaming      & 715   & 1,224 & 3,537 & 6,539           \\ \midrule
drugs             & 3,870 & 3,858 & 8,618 & 5,717          \\ \midrule
gambling          & 1,393 & 2,058 & 2,693 & 6,250          \\ \midrule
health shaming    & 1,744 & 1,737 & 7,270 & 6,903          \\ \midrule
offline crime     & 1,037 & 1,806 & 2,206 & 5,858           \\ \midrule
online crime      & 1,058 & 990 & 3,181 & 5,455           \\ \midrule
politics          & 1,593 & 2,131 & 7,650 & 6,150           \\ \midrule
pornography       & 1,289 & 2,048 & 2,824 & 5,775           \\ \midrule
prostitution      & 634   & 1,289 & 240   & 5,990          \\ \midrule
racism            & 1,156 & 1,565 & 3,760 & 5,849           \\ \midrule
religion          & 4,110 & 4,102 & 2,869 & 5,520           \\ \midrule
sexism            & 1,970 & 1,600 & 754   & 5,899          \\ \midrule
sexual minorities & 1,022 & 1,941 & 3,644 & 5,130           \\ \midrule
slavery           & 288   & 1,045 & 442   & 4,538           \\ \midrule
social injustice  & 1,230 & 1,884 & 5,294 & 6,446           \\ \midrule
suicide           & 1,420 & 1,405 & 1,931 & 4,117           \\ \midrule
terrorism         & 577   & 1,295 & 310   & 4,470          \\ \midrule
weapons           & 1,530 & 2,226 & 726   & 5,546          \\ \bottomrule
\end{tabular}
\caption{Number of samples per topic in sensitive topics and inappropriateness datasets for both versions of collected datasets. We provide the statistics for the datasets collected during the first stage (v.1) described in~\cite{babakov-etal-2021-detecting} and for the new versions of the datasets collected later and reported here for the first time (v.2). For the new version of the inappropriateness dataset we also report the statistics for the fraction of the data which contains only samples annotated as appropriate or inappropriate with the confidence of 100\%.}
\label{tab:sensitive_statistics}
\end{table}

\subsection{Dataset of Sensitive Topics}
The dataset of sensitive topics consists of 33,904 unique samples (sentences). Table~\ref{tab:sensitive_statistics} shows the distribution of samples across topics and also compares the dataset with its old version described in~\cite{babakov-etal-2021-detecting}. We provide the examples of topic-labelled utterances in Appendix~\ref{section:appendix_A}. 

\begin{figure}[ht!]
\centering
\includegraphics[width=0.7\linewidth]{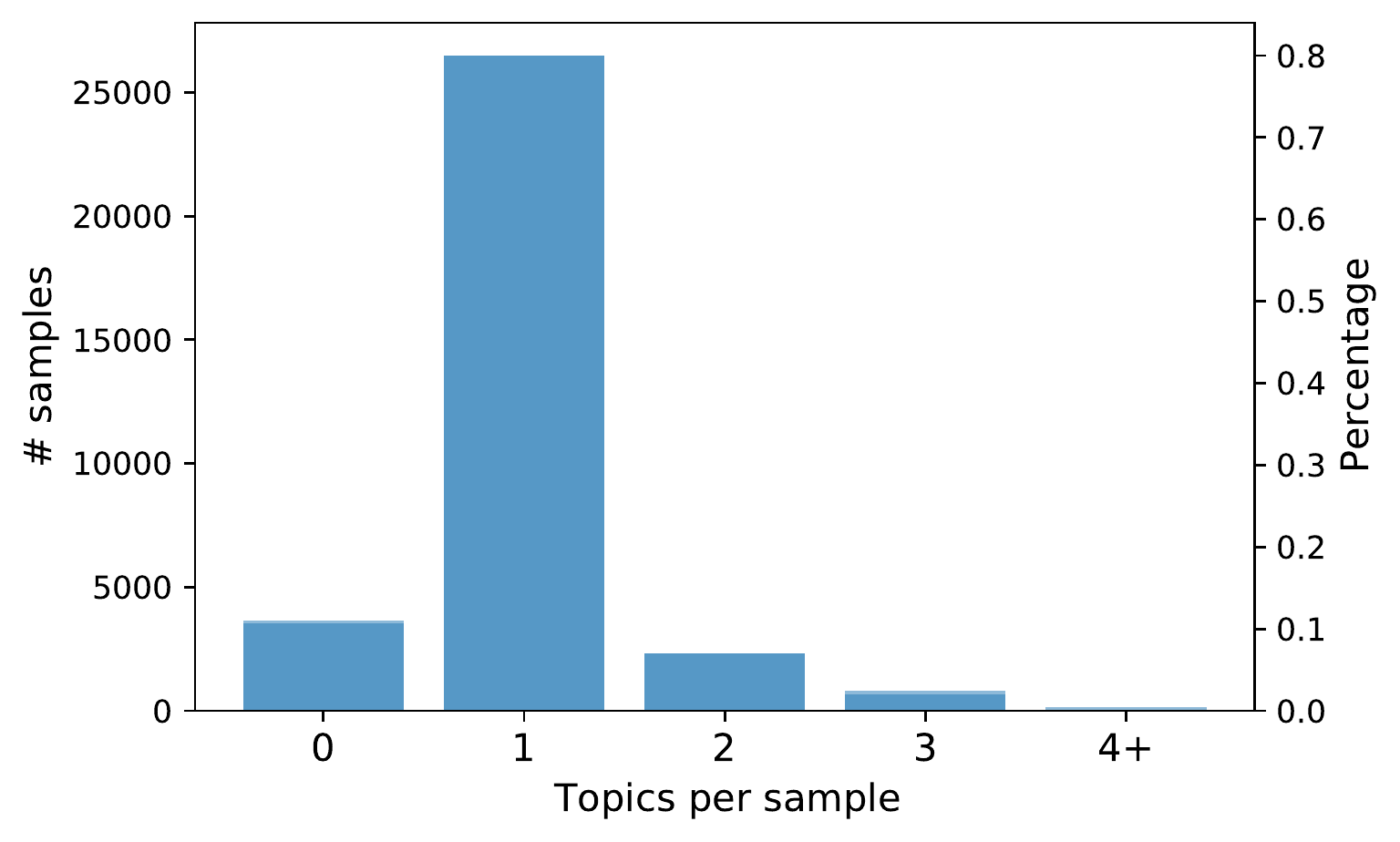}
\caption{Distribution of topics per sample samples in manually annotated part of sensitive topics dataset.}
\label{fig:topics_per_sample_chart}
\end{figure}

One sample can relate to more than one topic. Our analysis shows that there are 10\% of such examples in the manually annotated data (see Figure~\ref{fig:topics_per_sample_chart}). The co-occurrence of topics is not random. It indicates the intersection of multiple topics. The most common co-occurrences are ``politics, racism, social injustice'', ``prostitution, pornography'', ``sexual minorities, pornography''.  
In contrast, 13\% of samples in the topic dataset do not touch any sensitive topic. These are examples that were pre-selected for manual topic annotation using keywords and then were annotated as not related to the topics of interest. They were added to the dataset so that the classifier trained on this data does not rely solely on keywords.

\begin{figure}[ht!]
\centering
\includegraphics[width=0.7\linewidth]{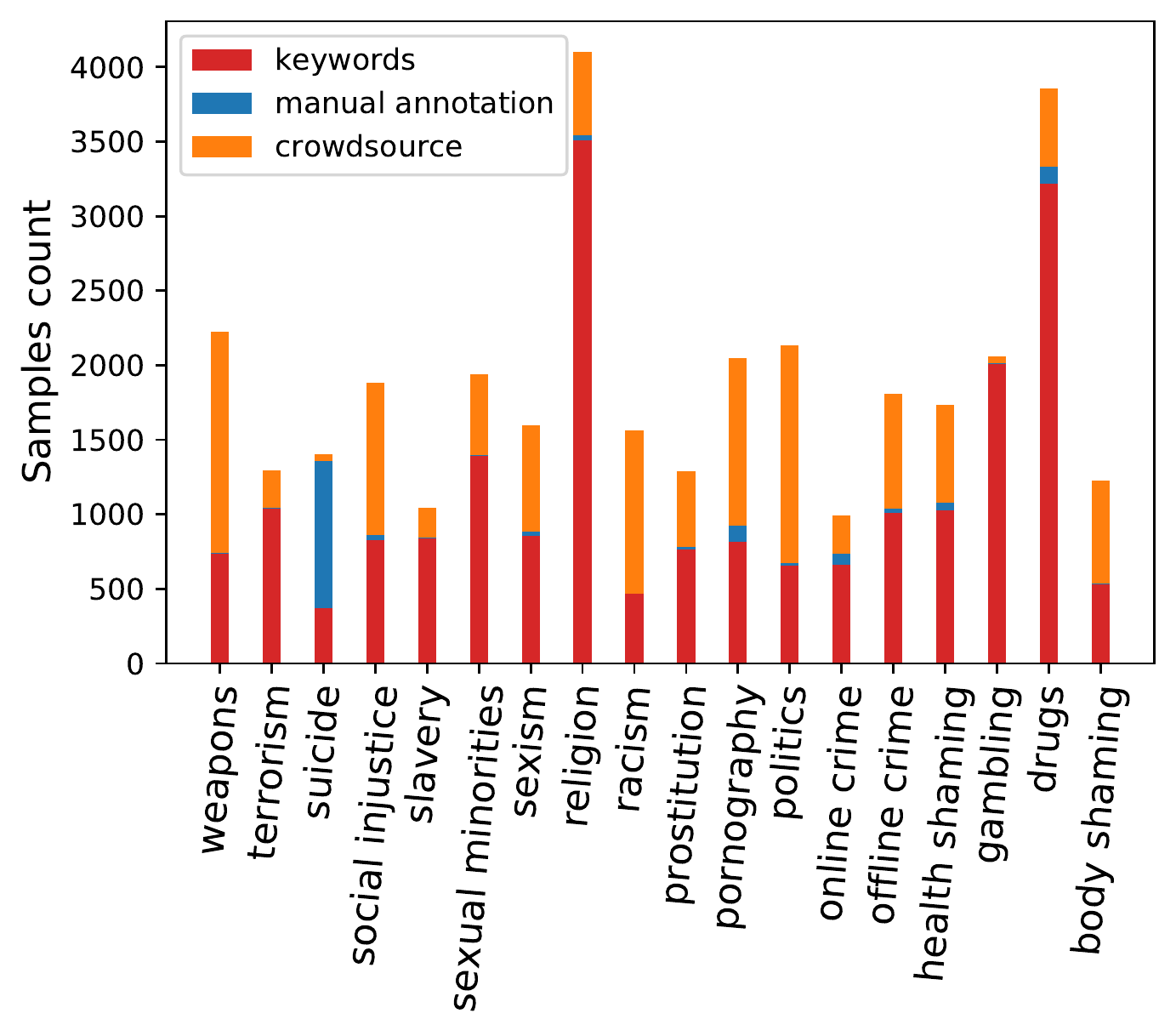}
\caption{Statistics of sources of the samples for sensitive topics dataset. \textit{keywords} -- data collected via inherent keywords search, \textit{manual annotation} -- data annotated by experts from our team, \textit{crowdsource} -- data annotated by crowd workers.}
\label{fig:multilabel_source_plot}
\end{figure}

As described in Section~\ref{section:topic_labeling}, we used three labelling strategies: crowd annotation, expert annotation, and automatic keyword-based annotation. The number of samples of different topics annotated via the three strategies is shown in Figure~\ref{fig:multilabel_source_plot}. 

    \paragraph{Data labelled via crowdsourcing} 9,670 samples were annotated via crowdsourcing. The average number of annotations per example is 4.3; the average time to label one example is 10.8 seconds. This part of sensitive topics datasets was annotated by 926 unique crowd workers. We admitted only workers who are native speakers of Russian.

    We calculate the inter-annotator agreement using Krippendorf's alpha~\cite{krippendorf} coefficient which is one minus the ratio of observed and expected disagreement. Since our topic dataset is multilabel (i.e. a sample can be assigned any number of the topic labels), we use MASI distance~\cite{Passonneau2006MeasuringAO} to measure the distance between the sets of labels assigned to a sample by different annotators. MASI is defined as a multiplication of Jaccard index $J$ and monotonicity index $M$, where $J=\frac{P \cap S}{P \cup S}$ and $M=1$ for equal sets, $M=2/3$ if one of the sets is a subset of the other, $M=1/3$ for intersecting sets none of which is a subset of the other, and $M=0$ if the sets intersection is empty.
    
    When considering all crowdsourced samples without filtering, the Krippendorff's alpha is 0.35 which corresponds to low agreement. Possible reasons for this result are the large number of labels, ambiguous samples or workers who did not understand the task or cheated. 
    To eliminate the two latter problems, we filter the annotated data. We remove the samples for which the Dawid-Skene labelling confidence is below 0.98 and recompute Krippendorff's alpha for the rest of data. The new value of the inter-annotator agreement measure is 0.66, which is reasonable agreement. We filter out 904 (9\%) samples. We assume that these samples are difficult for crowd workers to annotate, so we re-label them manually.

    We also compute the agreement between the crowd workers and experts. We use only the filtered sentences. To get expert annotations, we randomly sample 10 utterances from each topic, yielding 180 unique utterances and annotate them ourselves. Krippendorff's alpha agreement between three experts is 0.7. The expert annotations are then aggregated via majority voting (a sample is considered to belong to a topic if 2 out of 3 experts agreed on this topic), the aggregated labels are considered gold standard. The Krippendorff's alpha between the expert and crowd annotations is 0.72. The inter-annotator agreement for different subsets of the dataset is shown in Table~\ref{tab:multilabel_agreement}.
  
    \paragraph{Data from keyword search} 21,656 samples were collected by inherent keywords search on specialized forums. Analogously to crowdsourced labelling, here we also calculate the agreement of keyword-based and expert annotations. We randomly sample 10 utterances related to each topic and get 178 unique utterances. These texts are manually annotated by three experts, the Krippendorff's alpha between them is 0.73. We then aggregate their annotations as described above and compute the Krippendorff's alpha agreement of the expert and keyword-based annotations. This experiment yields Krippendorff's alpha of 0.76.
    
    \paragraph{Manual annotation by our team} 2,578 samples were manually labeled solely by the experts. These samples include both verification of samples labelled with the pre-trained multilabel classifier (Section~\ref{multilable_classifier_relabeling}) and annotation of samples fetched from the resources dedicated to a particular topic (Section~\ref{inherent_kw}). The Krippendorff's alpha of the three experts for this data is 0.68.

\begin{table}[]
\centering
\begin{tabular}{l|c|c|c||c|c}
\toprule
\multicolumn{2}{c|}{\textbf{Data collection approach}} & \multicolumn{1}{l}{\textbf{Initial data}} & \multicolumn{1}{|l||}{\begin{tabular}[c]{@{}l@{}}\textbf{Filtered}\\ \textbf{data}\end{tabular}} & \multicolumn{1}{l|}{\begin{tabular}[c]{@{}l@{}} \textbf{Experts}\\ \textbf{IAA}\end{tabular}} & \multicolumn{1}{l}{\begin{tabular}[c]{@{}l@{}}\textbf{Experts vs}\\ \textbf{other IAA}\end{tabular}} \\ \midrule
\multirow{2}{*}{\begin{tabular}[c]{@{}l@{}}crowd-\\sourcing\end{tabular}} & \# samples & 9,670 & 8,766 & \multicolumn{2}{c}{180} \\ \cmidrule{2-6} 
 & agreement & 0.35 & 0.66 & 0.70 & 0.72 \\ \midrule
\multirow{2}{*}{keywords} & \# samples & 21,656 & -- & \multicolumn{2}{c}{178} \\ \cmidrule{2-6} 
 & agreement & -- & --  & 0.73 & 0.76 \\ \midrule
\multirow{2}{*}{\begin{tabular}[c]{@{}l@{}}expert\\ annotation\end{tabular}} & \# samples & 2,578  & -- & -- & -- \\ \cmidrule{2-6} 
 & agreement & 0.68 & -- & -- & -- \\ \bottomrule
\end{tabular}
\caption{Krippendorff's alpha inter-annotator agreement (IAA) coefficient for different subsets of the sensitive topics dataset. \textit{Initial data} --- the number of samples collected in different annotation setups. \textit{Filtered data} --- the number of samples labelled by crowd workers with the high confidence. \textit{Experts IAA} --- the agreement between the experts within the selected data fraction, \textit{Experts vs other IAA} --- the agreement between the aggregated expert annotations and other annotation strategy (crowdsourced or keyword-based).}
\label{tab:multilabel_agreement}
\end{table}

\subsection{Inappropriateness Dataset}

The originally collected inappropriateness dataset consists of 163,332 unique samples. 8,822 of these samples also belong to the sensitive topics dataset and thus have manually assigned topic labels. The other 154,510 samples have topic labels defined automatically using a BERT-based sensitive topics classification model (described in Section~\ref{sec:benchmarking}). 
The average number of annotations per example is 3.5; the average time to label one example is 7 seconds. The inappropriateness datasets was annotated by 2,680 unique crowdsource workers. Analogously to the sensitive topics dataset, here we hired only Russian native speakers. 

We provide the samples of appropriate and inappropriate utterances on sensitive topics in Appendix~\ref{section:appendix_B}. 

For this dataset we also compute inter-annotator agreement as the Krippendorff's alpha coefficient. Krippendorff's alpha for the whole dataset is low -- 0.39. Analogously to the sensitive topics dataset, we filter the inappropriateness corpus to remove annotations by malicious or low-performing users. The filtering consists of two stages. First, we remove the samples with the low Dawid-Skene confidence. We test three thresholds: 0.8, 0.9, and 0.95. We also filter the samples by the number of votes for and against the presence of inappropriateness in a sample. We consider two filters: (i)~\textbf{all} -- the samples where all workers agreed on the sample label (appropriate or inappropriate) and (ii)~\textbf{all-but-one} -- all workers but one agreed on the sample label. 

The optimal balance of the kept data and the agreement is achieved if we filter out sentences with with the Dawid-Skene confidence of below 0.9 where two or more workers voted against the final label. This leaves us with 124,597 samples which is 76\% of the initial dataset and the inter-annotator agreement is 0.65 which corresponds to reasonable agreement. However, we release the whole big dataset, so it can be used as is or re-labelled to validate the low-confidence labels. The sizes of the corpus subsets yielded by different filtering approaches and their Krippendorff's alpha scores are given in Table~\ref{tab:inapp_filtering}.

\begin{table}[]
\centering
\begin{tabular}{c|c|c|c}
\toprule
\multicolumn{1}{l|}{\begin{tabular}[c]{@{}l@{}}\textbf{Dawid-Skene} \textbf{confidence}\end{tabular}} & \textbf{Votes distribution} & \multicolumn{1}{l|}{\textbf{\# Samples}} & \multicolumn{1}{l}{\begin{tabular}[c]{@{}l@{}}\textbf{Krippendorff's} $\alpha$\end{tabular}} \\ \midrule
\multirow{3}{*}{$\geq$~0.85} & all & 99,197 & 1 \\ \cmidrule{2-4} 
 & all-but-one & 132,904 & 0.60\\ \cmidrule{2-4} 
 & no filtering & 138,160 & 0.55 \\ \midrule
\multirow{3}{*}{$\geq$~0.90} & all & 97,310 & 1 \\ \cmidrule{2-4} 
 & all-but-one & 124,597 & 0.65 \\ \cmidrule{2-4} 
 & no filtering & 129,178 & 0.60 \\ \midrule
\multirow{3}{*}{$\geq$~0.95} & all & 90,383 & 1 \\ \cmidrule{2-4} 
 & all-but-one & 108,991 & 0.72 \\ \cmidrule{2-4} 
 & no filtering & 112,093 & 0.68\\ \bottomrule
\end{tabular}
\caption{Application of various filtering approaches to the dataset of inappropriate utterances. \textbf{all} -- samples where all workers agreed on the sample label (appropriate or inappropriate), \textbf{all-but-one} -- all workers but one agreed on the sample label.} \label{tab:inapp_filtering}
\end{table}

Analogously to the sensitive topics dataset, we compute the inter-annotator agreement between the crowdsourced and the expert annotation. Cohen's kappa agreement coefficient~\cite{doi:10.1177/001316446002000104} computed for two expert annotators (authors of the work) on 180 sentences is 0.66. Cohen's kappa between the averaged expert labelling and the aggregated crowd workers labelling (on the filtered dataset) is 0.72.

\begin{figure}[]
\centering
\includegraphics[scale=0.4]{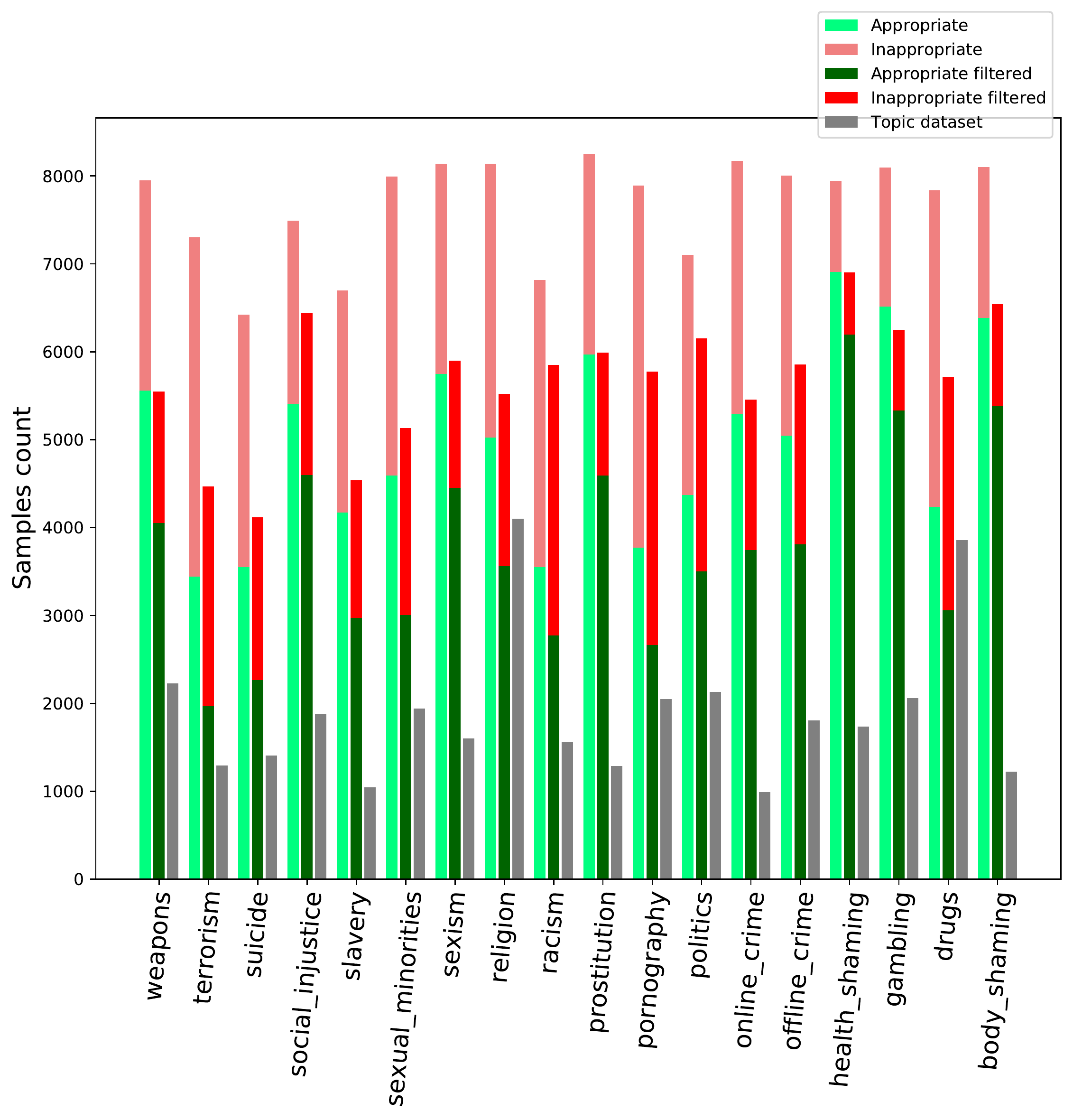}
\caption{\textit{Left bar}: number of appropriate (light green) and inappropriate (coral) samples per topic in the originally collected inappropriateness dataset. \textit{Central bar}: number of appropriate (dark green) and inappropriate (red) samples per topic in the final version of the dataset filtered by confidence and agreement. \textit{Right bar}: number of samples per topic in the dataset of sensitive topics.}
\label{fig:all_datasets_counts}
\end{figure}

We compare the full dataset with its filtered version to make sure they are homogeneous. First, we visualize the distribution of appropriate and inappropriate samples within each topic in Figure~\ref{fig:all_datasets_counts}. It shows that the balance of all topics and the proportion of appropriate and inappropriate utterances in the original and the filtered datasets are the close. To validate this visual intuition we calculate the average percentage of appropriate samples per each topic for both version of datasets. The results are 0.643 and 0.655 for the original and the filtered versions, respectively. We also check if the decrease in the number of samples was uniform across different topics. We calculate the percentage of removed samples per topic. The average percentage  is 26\% with the standard deviation of 0.08.

\begin{figure}[]
\centering
\includegraphics[scale=0.4]{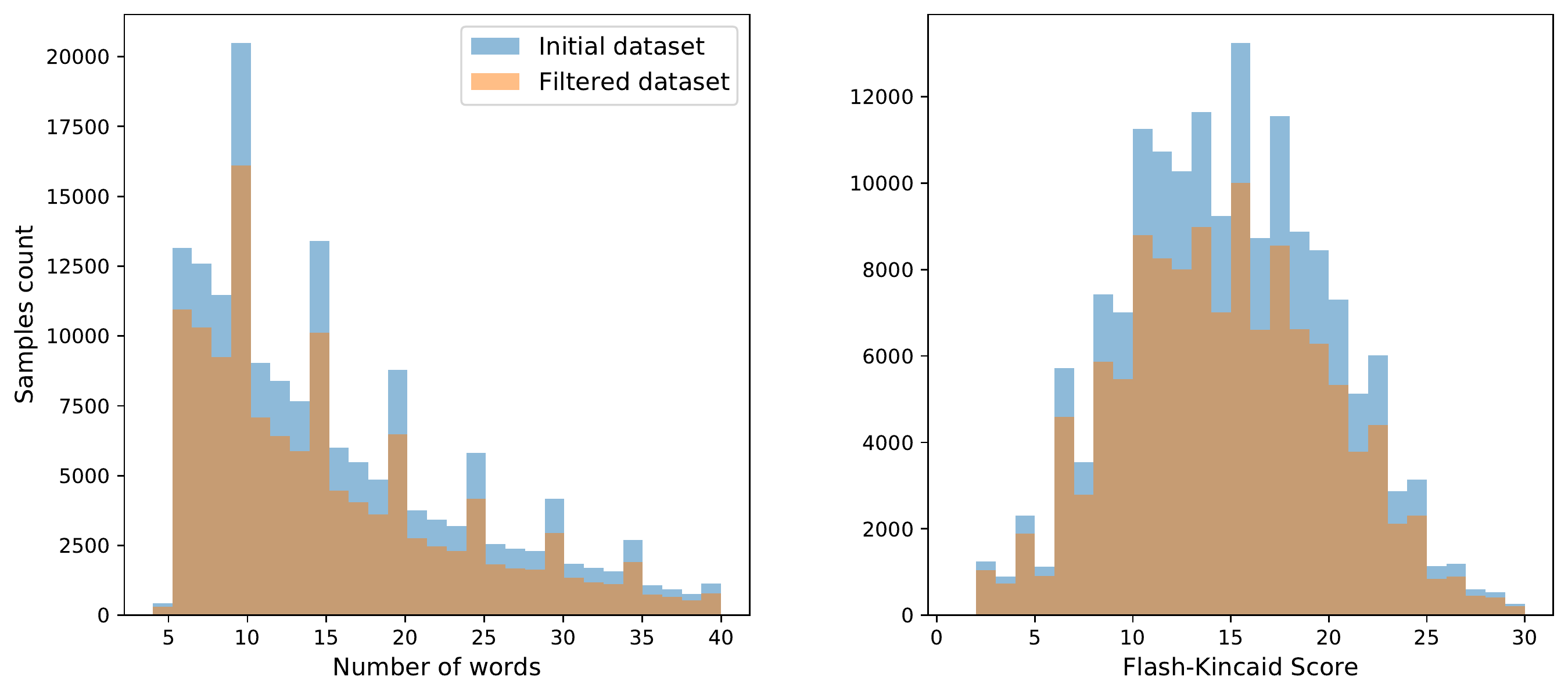}
\caption{The comparison of words count and Flesch–Kincaid readability score distribution between originally collected and filtered datasets}
\label{fig:filtered_and_original_comparison}
\end{figure}

Another characteristic of a corpus is the readability of its samples. It could have happened that the sentences with the low confidence were more difficult, and our filtering removed these more complicated samples. To check the readability of the original and the filtered versions of the inappropriateness datasets, we plot the the distribution of number of words and the Flesch–Kincaid readability score~\cite{flesch1979write} for them in Figure~\ref{fig:filtered_and_original_comparison}. The shapes of the distributions do not change after the filtering, so the readability characteristics of the dataset stay the same.

\subsection{Common datasets statistics}

Table~\ref{tab:sensitive_statistics} shows the number of samples on each sensitive topic in both datasets. We also show the number of appropriate and inappropriate samples within each topic and number of samples collected for each topic in sensitive topics dataset in Figure~\ref{fig:all_datasets_counts}.

The samples in the datasets are mostly single sentences; their average length is 16 words for the inappropriateness dataset and 18 words for the topic dataset. The sample length for different topics ranges from 14 to 21 words. 

\section{Results and Discussion}
\label{sec:benchmarking}

In this section, we showcase the usefulness of the collected data by training classification models on both datasets. We fine-tune pre-trained ruBERT model (a BERT model trained on Russian texts \cite{kuratov2019adaptation}) and a number of other baseline models on our data.
    We use \texttt{transformers} library\footnote{\url{https://huggingface.co/transformers}} with pre-trained Conversational RuBERT\footnote{\url{https://huggingface.co/DeepPavlov/rubert-base-cased-conversational}} weights.

\subsection{Classifier of Sensitive Topics}
\label{sec:topic_classifier}

We train a sensitive topics classifier on all automatically labelled data and on 70\% of manually annotated samples from sensitive topics dataset. We use the rest of the manually annotated data for tuning and testing of the model: 1,700 samples are used as the validation set and 1,700 are used as test set. We make sure to validate and test the model only on manually annotated data which is more diverse and more accurate. This is necessary since a large part of the dataset was labelled automatically via keyword search. Thus, using keywords-labelled samples for testing sets may yield a too simple test  this will most probably yield exaggerated classification metrics.

\begin{figure}
\centering
\includegraphics[width=0.8\linewidth]{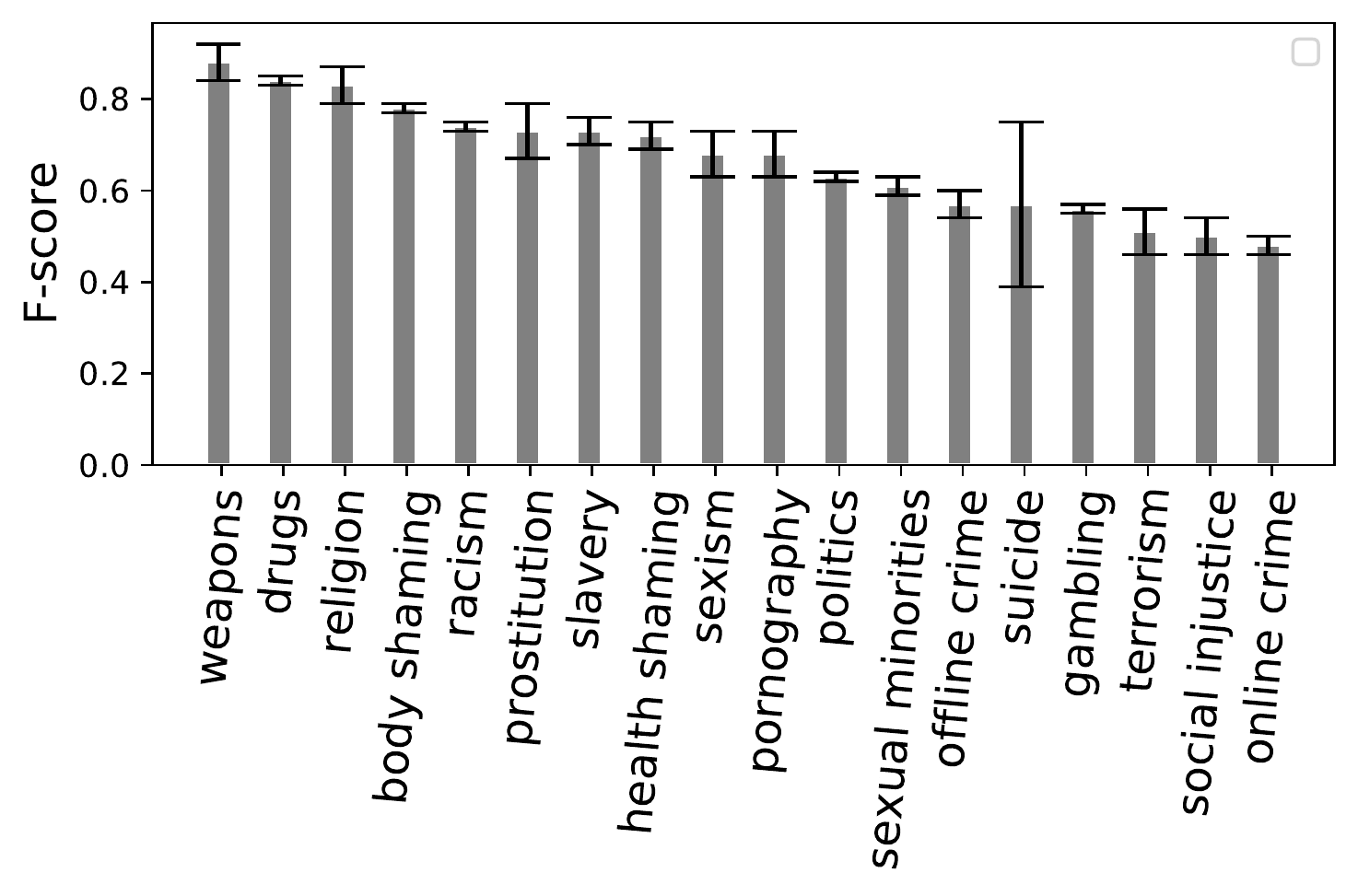}
\caption{F$_1$-scores of the BERT-based sensitive topics classifier.}
\label{fig:chart}
\end{figure}

\begin{figure}
\centering
\includegraphics[scale=0.7]{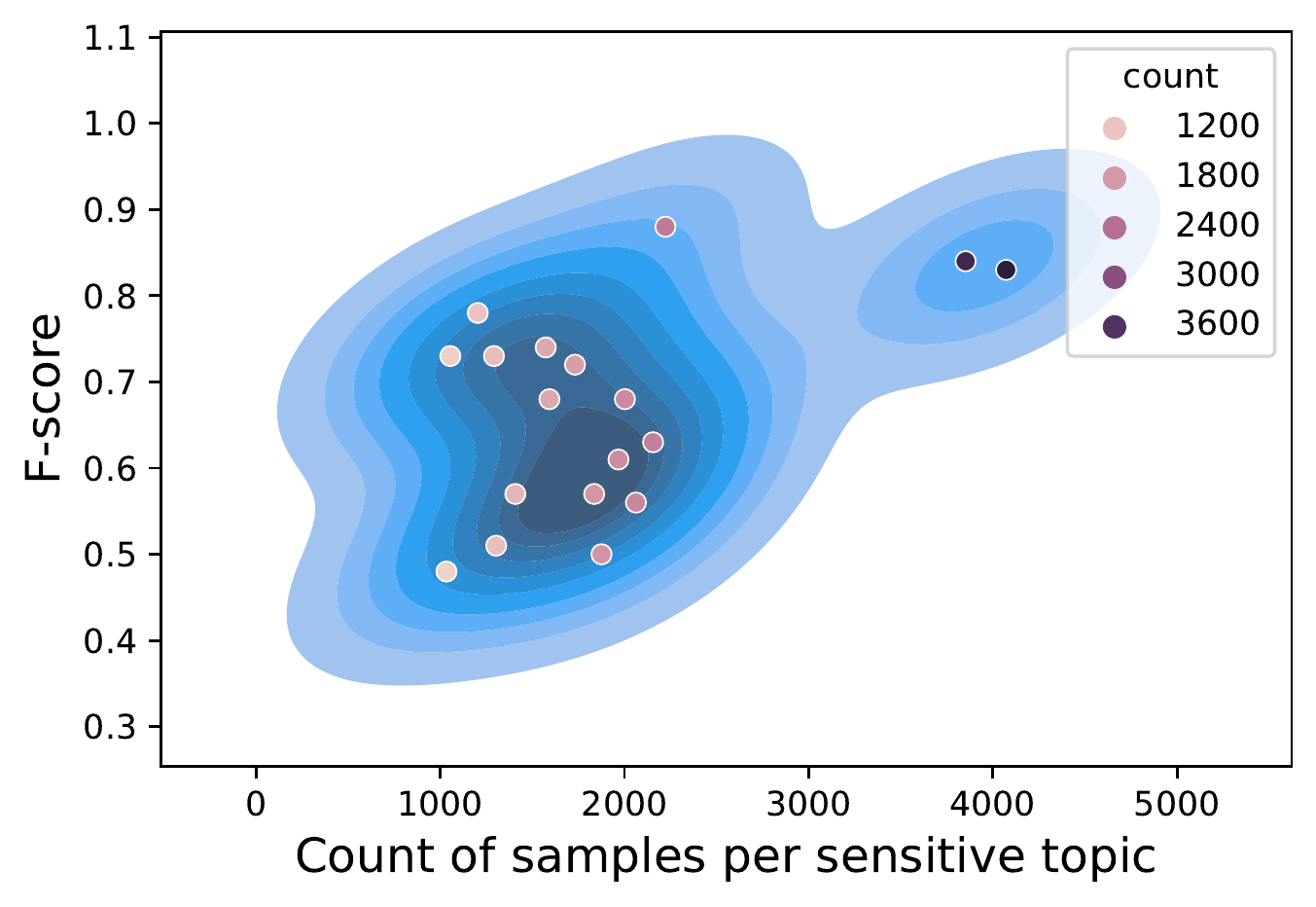}
\caption{Relationship between the number of samples per sensitive topic and the F$_1$-score of the classifier for this topic.}
\label{fig:multi_count_vs_fscore}
\end{figure}

We measure the  performance of different classifiers with the precision, recall, F$_1$-score and ROC-AUC (see Table~\ref{tab:multilabel_metrics}). We perform 5-fold cross-validation. The F$_1$-scores for individual topics from the best-performing conversational ruBERT classifier are shown in Figure~\ref{fig:chart}. The scores of individual topics are correlated with the number of training examples (see Figure~\ref{fig:multi_count_vs_fscore}). However, some topics (e.g. politics) have a low score despite being well represented in the data. This probably indicates the fact that the topic is complex and can be separated into several subtopics, such as: military service, pure political agenda (elections, referendums, meetings), historical and contemporary conflicts.

\begin{table}[]
\centering
\begin{tabular}{l|c|c|c|c}
\toprule
\textbf{Model} & \multicolumn{1}{l|}{\textbf{Precision}} & \multicolumn{1}{l|}{\textbf{Recall}} & \multicolumn{1}{l|}{\textbf{F-score}} & \multicolumn{1}{l}{\textbf{ROC-AUC}} \\ \midrule
\multicolumn{5}{c}{Baselines} \\ \midrule
\begin{tabular}[c]{@{}l@{}}Logistic Regression +\\ TF-IDF\end{tabular}  & \textbf{0.90}$\pm$0.02 & 0.57$\pm$0.01 & 0.60$\pm$0.01 & \textbf{0.90}$\pm$0.02 \\ \midrule
Naive Bayes + TF-IDF & 0.59$\pm$0.02 & 0.51$\pm$0.02 & 0.51$\pm$0.01 & 0.79$\pm$0.02 \\ \midrule
SVM + TF-IDF & 0.46$\pm$0.01 & 0.50$\pm$0.01 & 0.48$\pm$0.02 & \textbf{0.88}$\pm$0.01 \\ \midrule
CNN + fastText & 0.75$\pm$0.03 & 0.58$\pm$0.02 & 0.64$\pm$0.01 & 0.80$\pm$0.02 \\ \midrule
\multicolumn{5}{c}{BERT} \\ \midrule
Conversational ruBERT & 0.76$\pm$0.01 & \textbf{0.67}$\pm$0.03 & \textbf{0.71}$\pm$0.02 & 0.82$\pm$0.01 \\ \bottomrule
\end{tabular}
\caption{Performance of topical classifiers trained on the sensitive topics dataset.}
\label{tab:multilabel_metrics}
\end{table}

\subsection{Inappropriateness Classifier}
\label{sec:binary_classifier}

Analogously to the sensitive topics classifier, we train the inappropriateness classifier on 80\% of the inappropriateness-annotated messages, and use 10\% for validation and the rest 10\% for testing. Our main model is ruBERT fine-tuned on the inappropriateness dataset. 

In addition to this model, we also test other approaches. 
We train three baselines: a Logistic Regression classifier, a Naive Bayes classifier, and a Stochastic Gradient Descent Classifier (\textit{SGD}). All of them use TF-IDF vectors of lemmatized texts of samples as features.
We also test other approaches based on neural networks. Namely, we train a convolutional neural network (CNN) \cite{NIPS1989_53c3bce6} which uses pre-trained word vectors as input. We experiment with fastText \cite{bojanowski2017enriching} vectors. Table~\ref{tab:hybrid_scores} shows the performance of all trained models.

\subsection{Use of Topical Information for Inappropriateness Classification}
\label{sec:topic_dependent_classification}

We also infuse the information about topics into the inappropriateness classifier in order to further improve its performance. We perform two lines of experiments: (1)~add topic-dependent layer to the output and (2)~add topic information to the BERT input.

\textit{Topic-specific layer in the BERT output.} We initialize topic embeddings randomly, multiply the BERT output by the embedding of a corresponding topic, and pass the result to the softmax output layer. 
This additional topic layer has no effect on the classifier performance: the classifier enhanced with these topic embeddings shows the same F$_1$-score as the original classifier.

\textit{Special input to BERT.} We use the approach similar to the one described in \cite{8903313}. We extend BERT vocabulary with 18 special tokens corresponding to sensitive topics and one to indicate the absence thereof. Based on the topic(s) of a sample we add the corresponding topic tokens to the input and separate them from the text data using the [SEP] token. To better separate the topic tokens from the text tokens, we use BERT segment embeddings: the sequence of word tokens is considered sentence A, and the sequence of topic tokens is considered sentence B.

We employ different approaches for initialization of topic embeddings:
\begin{itemize}
    \item We initialize the embeddings randomly.
    \item We assemble sentence vectors (the vectors of [CLS] token) of all samples with a particular topic and use their average or max-pooled vectors as topic embedding.
    \item We use the linear layer of the BERT head in the sensitive topics classifier described in Section~\ref{sec:topic_classifier}. For multi-topic samples we apply either max-pooling or averaging to combine multiple topic vectors to a single vector.
\end{itemize}

Our preliminary experiments showed that max-pooling of the linear layer from the sensitive topics classifier performs better than the other tested architectures, so we report its result here. We conduct 5-fold cross-validation to evaluate its stability. Table~\ref{tab:hybrid_scores} shows the performance of topic-enhanced classifier (bottom line) and gives the scores of the classifiers without topical information for comparison. Even though the results of topic-enhanced classification did show a small improvement of performance, the McNemar’s test returned the $p$-value of 0.25 showing that these improvements are not statistically significant. 
However, the performance of both standard and topic-enhanced inappropriateness classifiers based on conversational ruBERT is significantly better than that of the other approaches with the significance level $\alpha < 0.05$. 

\begin{table}[]
\centering
\begin{tabular}{l|c|c|c|c}
\toprule
\textbf{Model} & \textbf{Precision} & \textbf{Recall} & \textbf{F-score} & \textbf{ROC-AUC} \\ \midrule
\multicolumn{5}{c}{Baselines} \\ \midrule
Logistic Regression + TF-IDF & 0.77$\pm$0.01 & 0.78$\pm$0.02 & 0.78$\pm$0.02 & 0.79$\pm$0.01 \\ \midrule
Naive Bayes + TF-IDF & 0.78$\pm$0.02 & 0.75$\pm$0.01 & 0.66$\pm$0.03 & 0.77$\pm$0.01 \\ \midrule
SVM + TF-IDF & 0.53$\pm$0.02 & 0.73$\pm$0.01 & 0.62$\pm$0.02 & 0.71$\pm$0.02 \\ \midrule
CNN + fastText & \textbf{0.85}$\pm$0.05 & 0.79$\pm$0.03 & 0.81$\pm$0.04 & 0.88$\pm$0.03 \\ \midrule
\multicolumn{5}{c}{BERT-based classifiers} \\ \midrule
Conversational ruBERT & \textbf{0.87}$\pm$0.01 & \textbf{0.87}$\pm$0.02 & \textbf{0.87}$\pm$0.01 & \textbf{0.93}$\pm$0.01 \\ \midrule
Conversational ruBERT + topic & \textbf{0.88}$\pm$0.02 & \textbf{0.88}$\pm$0.01 & \textbf{0.88}$\pm$0.01 & \textbf{0.93}$\pm$0.02 \\ \bottomrule
\end{tabular}
\caption{Performance of different binary inappropriateness classifiers. The best-performing ruBERT-based model was used for testing the usefulness of the topic information for the improvement of classification.}
\label{tab:hybrid_scores}
\end{table}

\subsection{Manual evaluation of inappropriateness filtering}

The main goal of our work is to make sure that the information which can harm the reputation of a company but cannot be detected with general toxicity classifiers will not reach the final user. 

To test whether the inappropriateness classifier trained on the collected dataset is capable of accomplishing this task, we perform the following experiment. We aim at discovering if the inappropriateness classifier can detect inappropriate samples which were recognized as non-toxic.
We perform the following stages:

\begin{itemize}
    \item We fetch random utterances from \href{https://2ch.hk}{2ch.hk} which is not moderated so is likely to contain toxicity and inappropriateness.
    \item We filter out utterances which contain obscene words.
    \item We classify the remaining utterances with the toxicity classifier (same as used in Section~\ref{sec:annot}) and mark them as \textit{toxic} if they got the confidence of 0.75 or above.
    \item We classify these utterances with our best-performing inappropriateness classifier and mark them as \textit{inappropriate} if they got the confidence of 0.75 or above.
    \item We also classify all utterances with the topic classifier.
\end{itemize}

Thus, the utterances can have one of four combinations of inappropriateness and toxicity labels: (i)~toxic and inappropriate, (ii)~toxic and appropriate, (iii)~non-toxic and inappropriate, (iv)~non-toxic and appropriate. In addition to that, each of them can be attributed to one or more sensitive topics or have no topic.

To check the performance of the classifiers, we conduct manual evaluation. For the manual evaluation we randomly select sentences from this automatically labelled data. We select 100 sentences which belong to at least one sensitive topic so that there are 25 examples of each toxicity+inappropriateness combinations. Then we select 100 sentences which do not touch any sensitive topics using the same principle.

We ask two experts annotate these 200 samples. We ask them to indicate samples which could be anyhow unacceptable in a respectful conversation or as an answer of a chatbot. Thus, in this case we ask to identify both inappropriateness and toxicity. We then aggregate their answers in a ``pessimistic'' way. Namely, an utterance is considered inappropriate if at least one expert indicated it as such.

The Cohen's kappa inter-annotator agreement between the annotators is 0.65 for utterances related to a sensitive topic and 0.72 for non-topical utterances. We also compute the precision of our inappropriateness classifier for two groups of samples: samples which are non-toxic \textit{but} inappropriate, and samples which are toxic and either appropriate or not.
The former samples are of greater importance for us, because they cannot be detected by a toxicity classifier but need to be filtered out. The latter samples are less important, because they are filtered out by the toxicity classifier.

Table~\ref{tab:manual_eval} shows that non-toxic+inappropriate sentences are detected with a high precision. This means that our inappropriateness classifier can be successfully used as a means of additional filtering of unwanted content.

\begin{table}[]
\centering
\begin{tabular}{c|c|c|c}
\toprule
\multirow{2}{*}{\textbf{Dataset}} & \multirow{2}{*}{\textbf{Cohen’s kappa}} & \multicolumn{2}{c}{\textbf{Precision}} \\ \cmidrule{3-4} 
 &  & \textit{non-toxic\&inapp} & \textit{toxic\&other} \\ \midrule
Sensitive topic & 0.65 & 0.88 & 0.88 \\ \midrule
No topic & 0.72 & 0.80 & 0.94 \\ \bottomrule
\end{tabular}
\caption{Results of the manual evaluation of the pipeline of toxicity and inappropriateness classifiers. We report the precision separately for topic-related and non-topical sentences and for (i)~non-toxic and inappropriate utterances (\textit{non-toxic\&inapp}) and (ii)~toxic and either appropriate or inappropriate utterances (\textit{toxic\&other})}
\label{tab:manual_eval}
\end{table}

\subsection{The Use-Case of the Pre-trained Classifiers}

\begin{figure}
\centering
\includegraphics[width=1.0\linewidth]{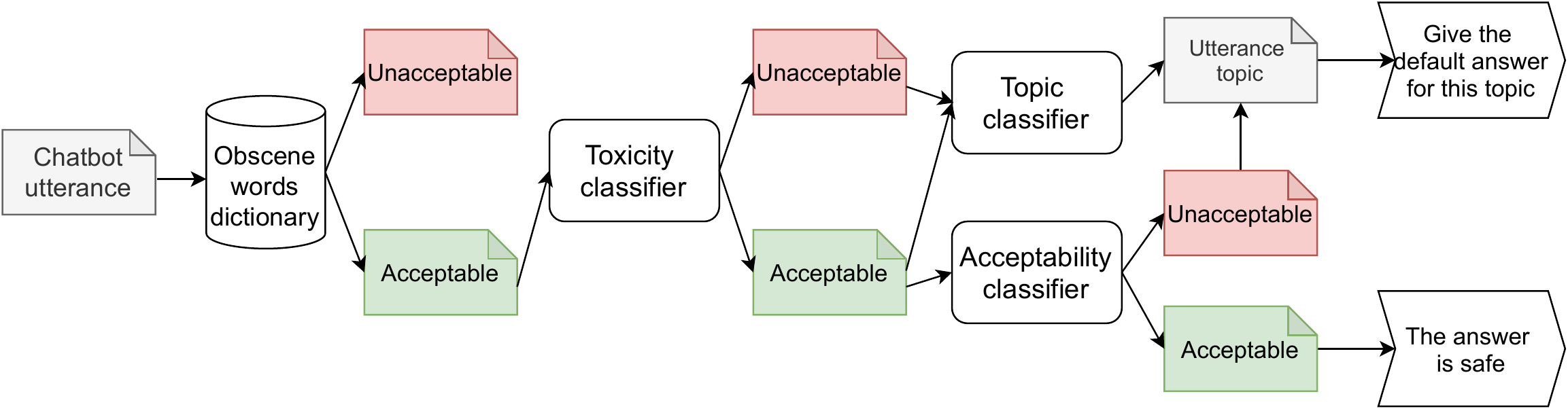}
\caption{Suggested usecase for the classifiers trained on the collected data.}
\label{fig:proposed_usage}
\end{figure}

The classifiers trained on our datasets could be used in the real-world tasks, such as filtering of utterances generated by an open-domain dialogue agent. We illustrate the idea in Figure~\ref{fig:proposed_usage}.

The utterances generated by a dialogue agent should first be checked for the presence of strong language using a dictionary of obscene words. Then, they should be checked for toxicity with a toxicity classifier. Finally, we can apply an inappropriateness classifier trained on our inappropriateness dataset. If an utterance is classified as toxic or inappropriate we can also define if it is related to any sensitive topic. Then the dialogue agent can replace the generated utterance with a pre-defined template answer for this topic, such as ``I don't feel like discussing terrorism'' in case the originally generated utterance is toxic or inappropriate and is also related to terrorism topic. 

Currently, a similar setup is being tested under by our partners at MTS. Also other large Russian internet companies approached us and testing the classifiers in their text content processing systems.

\section{Conclusions}

Toxicity in online communication is an acknowledged problem and its detection is a topic of active research.
In this work, we go beyond explicit toxicity and present an extended view of unwanted content in online communication. We notice that discussions on some \textit{sensitive} topics can cause frustration of readers and/or harm the reputation of a speaker. At the same time, not any discussion on these topics is unacceptable, so there exist \textit{appropriate} and \textit{inappropriate} utterances within each topic.

We introduce the task of detecting such inappropriate utterances. We form a list of sensitive topics which we target and collect a corpus of samples on these topics manually annotated for inappropriateness. In addition to that, we collect the corpus of samples manually annotated for the presence of sensitive topics. Both datasets contain texts in Russian. While we do not define inappropriateness explicitly but only provide annotators with examples of appropriate and inappropriate sentences, we see that the majority of them understand the notion of inappropriateness uniformly. This suggests that the inappropriateness has features which can be detected by machine learning models.

To check the usefulness of our datasets, we train two classification models on them: a multilabel sensitive topics classifier and a binary classifier of inappropriateness. We also make an attempt to incorporate topical information into the inappropriateness classification. However, this does not give any significant improvement, which is consistent with similar studies on context-aware toxicity classification. We release the collected corpora and models trained on them for non-commercial use.

\section{Future work}

First of all, the presented set of flammable/sensetive topics can be further extended covering more domains. Secondly, the inappropriateness classifier can be improved by incorporating the additional information: the level of politeness and formality of the utterance, the semantic roles in the text, etc. Finally, prominent direction of future work is to transfer the notion of inappropriateness to other languages by fine-tuning cross-lingual models, such as XLM-R~\cite{DBLP:journals/corr/abs-1911-02116} on the collected datasets.  

\section{Acknowledgements}

This work was conducted under the framework of the joint Skoltech-MTS laboratory. We are grateful for MTS for permission to share the produced datasets and code. We thank Pavel Odintsev for creating two illustrations of our technology reproduced in this article.

\bibliographystyle{unsrt}
\bibliography{lre_unsafe}

\newpage

\begin{appendices}
\section{Inappropriateness Dataset}
\label{section:appendix_A}

This section contains examples from the inappropriateness dataset described in Section~\ref{section:inappropriateness_labeling}.

\begin{longtable}{p{0.41\linewidth}|p{0.41\linewidth}|p{0.07\linewidth}}
\toprule
Original text                                                               & Translated text                                                     & Inapp. \\ \midrule
\foreignlanguage{russian}{Хвтит звонить в полицию и жаловаться на пытки}                               & Stop calling police and complaining about tortures                     & yes           \\ \midrule
\foreignlanguage{russian}{Мне Аллах подсказывает убить маленькую девуочку и с её головой у метро кричать Господь ВЕЛИК!} &
  Allah tells me to kill a little girl and go with her head at the subway entrance shouting God is GREAT! &
  yes \\ \midrule
\foreignlanguage{russian}{Всех жирных баб надо травить}                                                & All fat women should be bullied                                    & yes           \\ \midrule
\foreignlanguage{russian}{У этой жирухи классическое музыкальное образование образование}              & This fat woman has a classical music education                      & yes           \\ \midrule
\foreignlanguage{russian}{У него прыщавая рожа. Фу как же противно на нео смотреть}             & He's got a pimply face. Really disgusting                     & yes           \\ \midrule
\foreignlanguage{russian}{Полные девушки могут прекрасно одеваться и в итоге выглядеть не хуже и даже лучше чем обычные стройняшки} &
  Plump girls can dress perfectly and eventually look no worse and even better than ordinary slim girls &
  no \\ \midrule
\foreignlanguage{russian}{Красиво одеваться можно даже не имея больших денег и красивой фигуры} &
  You can dress up prettily even without having much money and a beautiful figure &
  no \\ \midrule
\foreignlanguage{russian}{Процесс взлома уже пошёл, теперь его не остановить.}                         & The hacking process has already started, now it cannot be stopped. & yes           \\ \midrule
\foreignlanguage{russian}{Лучше скача с торрента}                                                      & It would be easier to download that from torrent                    & yes           \\ \midrule
\foreignlanguage{russian}{Зеркало сайта работает. Держи адрес}                                         & The site mirror is working. Here's the address                      & yes           \\ \midrule
\foreignlanguage{russian}{По кривой дороке нарктоы пошел парнеь, а ведь мог стать успешным} &
  Along the crooked path of drug addiction the guy finally went, although he had had a potential &
  no \\ \midrule
\foreignlanguage{russian}{Наркотики тебя до добра не доведут}                                          & Drugs mean trouble                                    & no            \\ \midrule
\foreignlanguage{russian}{Где купить добротный спайс в москве?}                                        & Where can I  buy good spice in Moscow?                                & yes           \\ \midrule
\foreignlanguage{russian}{Казино казино казино это музыка песнь и вино}                                & Casino is all about music song and wine                             & no            \\ \midrule
\foreignlanguage{russian}{Букмекерская контора 1хstv и 1хt одна организация?}                          & Are 1xstv and 1xt bookmaker the same organization?                           & no            \\ \midrule

\foreignlanguage{russian}{Ладно бы видного деятеля завалили а тут какого то ноунейм н** преступника} & Killing a noname n** is not a big deal                              & yes           \\ \midrule
\foreignlanguage{russian}{Ну он опять пошел н** морды разукрашивать}                            & So again he went out beating n** faces                                       & yes           \\ \midrule

\foreignlanguage{russian}{Надо ставить по крупному! Только в конторе 1икс бет!}                        & You should bet big! Only in 1xbet!                                 & yes           \\ \midrule

\foreignlanguage{russian}{Этот даун опять упал на ровном месте. Как он дожил вообще до своего возраста?}                        & This down again fell out of the blue. How did he even live up to his age?                                & yes           \\ \midrule

\foreignlanguage{russian}{как себя успокоить, если сердце разрывается от боли?}                        & how to calm yourself if your heart is bursting with pain?                                & no           \\ \midrule

\foreignlanguage{russian}{Никакой напряженности. И в Египте, и в Израиле сейчас совершенно обычная для этих стран обстановка.}  & No tension. Both in Egypt and in Israel, the situation is now completely normal for these countries. & no  \\ \midrule

\foreignlanguage{russian}{Японцы притесняют наши права на японию, Путин введи войска}  & The Japanese are oppressing our rights to Japan, Putin bring in troops . & yes  \\ \midrule

\foreignlanguage{russian}{Хотелось бы в общих чертах понять, о чем говорит иностранец.}  & I would like to understand in general terms what the foreigner is talking about countries. & no  \\ \midrule

\foreignlanguage{russian}{А я где то указывал,что люблю темнокожих?Или Вы как бы меня упрекнуть решили?}  & Did I mention somewhere that I like dark-skinned people?Or have you decided to reproach me, as it were? & yes  \\ \midrule

\foreignlanguage{russian}{Кто пользовался микрозаймами под материнский капитал на покупку жилья. правомерно ли это?}  & Who used microloans for maternity capital to buy housing. is it legitimate? & no  \\ \midrule

\foreignlanguage{russian}{В украинских маленьких пенсиях Порошенко и Ко виноваты !}  & Poroshenko and Co are to blame for Ukrainian small pensions ! & yes  \\ \midrule

\foreignlanguage{russian}{самый первый танк «Вездеход» сделаный А.А.Пороховщиковым}  & the very first tank "Vezdehod" made by A.A. Porokhovshchikov & no  \\ \midrule

\foreignlanguage{russian}{А для убийства лучше иметь нож.}  & And it's better to have a knife to kill. & yes  \\ \midrule

\foreignlanguage{russian}{Зарабатываю по 500 тыяч в месяц чисто на удаче, на ставках}                  & I earn 500 thousand rubles a month purely on luck, on bets          & yes           \\ \midrule

\foreignlanguage{russian}{Задачка для криминалистов. Суицид или убийство?))}                  & A task for criminologists. Suicide or murder?))         & no           \\ \midrule

\caption{Examples of appropriate and inappropriate utterances.}
\label{tab:appendix_inappropriate}
\end{longtable}

\section{Dataset of Sensitive Topics}
\label{section:appendix_B}

This section contains examples from the dataset of sensitive topics described in Section~\ref{section:topic_labeling}.

\begin{longtable}{p{0.39\linewidth}|p{0.39\linewidth}|p{0.12\linewidth}}
\toprule
Original text &
  Translated text &
  Sensitive topic \\ \midrule
\foreignlanguage{russian}{я сын таджикского наркоборона. поставляли герыч брежневу.Войска СССР в Афгане vs войска НАТЫ} &

  I'm a son of a Tajik drug lord. they supplied heroin to Brezhnev.Soviet troops in Afghanistan vs NATO's troops &
  offline crime, drugs \\ \midrule
  
  \foreignlanguage{russian}{Вот смотрю на них как будто не виновные глаза,хотят на свободу дальше убивать.} &

  Here I look at them as if innocent eyes, they want to continue killing for freedom. &
  offline crime \\ \midrule
  
    \foreignlanguage{russian}{Злостное уклонение от алиментов на секундочку уголовная статья. До года лишения свободы.} &

  Malicious evasion of alimony for a second is a criminal article. Up to a year in prison. &
  offline crime \\ \midrule
 
    \foreignlanguage{russian}{Даже воры так не живут сейчас.} &

  Even thieves don't live like that now. &
  offline crime \\ \midrule

\foreignlanguage{russian}{Шалаш взломает твою вебку и будет шантажировать тебя твоей дрочкой} &
  shalash will hack your webcam and blackmail you with your wank &
  online crime, pornography \\ \midrule
  
\foreignlanguage{russian}{Ищу наставника для хакерской деятельности,Все схватываю на лету писать в лс} &
  I am looking for a mentor for hacking, I grasp everything on the fly to write in the bos &
  online crime \\ \midrule
  
\foreignlanguage{russian}{Заказал взлом wt жены все прошло успешно, рекомендую} &
  I ordered the hacking of my wife's wt, everything went well, I recommend it &
  online crime \\ \midrule
  
\foreignlanguage{russian}{Не пить пиво. С рыбой. Вообще пиво сейчас это химия. А в рыбе соль. Грибы тоже сильно в соли переходят. И пуринов в них много.} &
  Don't drink beer with fish snacks. In general, there's nothing genuine in beer any more. And fish is full of salt. Mushrooms are also converted to salt in your body. And also contain much purines. &
  drugs \\ \midrule
  
\foreignlanguage{russian}{Хорошая табличка у тебя по сигаретам...прям мотивирует} &
  You have a good sign for cigarettes...straight motivates &
  drugs \\ \midrule
  
\foreignlanguage{russian}{Священник имеет кучу молодых любовниц, курит, пьет, употребляет наркотики, приходит обдолбанным, укуренным...} &
  A priest has a bunch of young lovers, smokes, drinks, uses drugs, comes to work being stoned... &
  drugs, religion \\ \midrule
\foreignlanguage{russian}{Какие у букмекеров ставки на выборы сша 2020 трамп байден кто фаворит} &
  What are the bookmakers ' bets on the US election 2020 trump biden who is the front-runner &
  gambling \\ \midrule
  
 \foreignlanguage{russian}{Лучший способ остаться в выигрыше, это не связываться с казино, только так вы его обыграет} &
  The best way to win is not to mess with the casino, the only way you will beat it &
  gambling \\ \midrule 

\foreignlanguage{russian}{Анон, который рассказывал про таек, ты скажи, ты их без презика трахал? И сколько самая дорогая шлюха тебе обошлась в переводе на рубли?} &
  Anon who talked about Thai girls, tell me, you fucked them without a condom? And how much did the most expensive whore cost you in rubles? &
  porno- graphy, prostitution \\ \midrule
  
 \foreignlanguage{russian}{Мне нравятся с волосами на лобке, я нормальный?} & I like pubic hair, am I normal? &
  porno- graphy, body shaming \\ \midrule
  
 \foreignlanguage{russian}{Я не рекомендую заниматься анальным сексом и там много проблем.} & I do not recommend anal sex and there are a lot of problems.&
  porno- graphy \\ \midrule

\foreignlanguage{russian}{Давай я твою мать/дочь куплю в рабыни.} &
  Let me buy your mother/daughter as a slave. &
  slavery \\ \midrule
  
\foreignlanguage{russian}{Мир всегда будет делиться на рабов и рабовладельцев} &
  The world will always be divided into slaves and slaveholders &
  slavery, social injustice \\ \midrule
  
\foreignlanguage{russian}{Ничего плохого. Так же как и в воровстве еды у нищих, изнасилованиях или доведении до самоубийства.} &
  That's not bad. Neither is stealing food from beggars, rape, or incitement to suicide. &
  suicide, social injustice\\ \midrule
  
\foreignlanguage{russian}{В чем же причина такого количества самоубийств у людей различных полов и возрастов?} &
  What is the reason for so many suicides among people of different genders and ages? &
  suicide\\ \midrule
  
\foreignlanguage{russian}{3 террориста-смертника Подорвали себя 5 взрывов. Что то не сходится.} &
  3 suicide bombers. Blew themselves up 5 blasts. That doesn't quite add up. &
  terrorism \\ \midrule
\foreignlanguage{russian}{А есть пистолеты но с перцем внутри?удар 5 м2 либо перцовые патроны на осу} &
  And there are guns but with pepper inside? To hit 5 m2 or pepper cartridges for wasp &
  weapon \\ \midrule
\foreignlanguage{russian}{Уродливая шлюха в ебучем кресле.} &
  An ugly whore in a fucking chair. &
  body shaming \\ \midrule
\foreignlanguage{russian}{это стенокардия, далее инфаркт. Нужно срочно к кардиологу и бросать курить.} &
  This is angina, it can be followed by a heart attack. You need to see a cardiologist right away and quit smoking. &
  health shaming \\ \bottomrule
\caption{Examples of manually annotated sensitive topics utterances.}
\label{tab:appendix_sensitive}
\end{longtable}

\end{appendices}

\end{document}